\documentclass[accepted]{uai2024} 
                        
\usepackage[american]{babel}

\usepackage{natbib} 
\usepackage{mathtools} 
\usepackage{booktabs} 
\usepackage{tikz} 

\usepackage{hyperref}       

\hypersetup{
colorlinks   = true, 
urlcolor     = blue, 
linkcolor    = blue, 
citecolor    = blue 
}

\usepackage{url}            
\usepackage{booktabs}       
\usepackage{amsfonts}       
\usepackage{nicefrac}       
\usepackage{microtype}      
\usepackage{xcolor}         

\usepackage{algorithm}
\usepackage{algorithmic}

\usepackage{amsmath}  
\usepackage{multirow}

\usepackage{url}            
\usepackage{booktabs}       
\usepackage{amsfonts}     
\usepackage{amsthm}     
\usepackage{nicefrac}       
\usepackage{microtype}      
\usepackage{xcolor}
\usepackage{tabularx}

\usepackage{graphicx} 
\usepackage{subfigure} 

\usepackage{titletoc}

\newcommand\DoToC{%
  \startcontents
  \printcontents{}{1}{\textbf{Table of Contents}\vskip3pt\hrule\vskip5pt}
  \vskip3pt\hrule\vskip5pt
}

\usepackage{amssymb}
\usepackage{amsmath}
\usepackage{amsfonts}
\usepackage{multirow}

\DeclareMathOperator{\argmin}{\arg\min}

\newtheorem{definition}{Definition}
\newtheorem{remark}{Remark}

\usepackage{caption}

\usepackage{wrapfig}

\usepackage{color}
\definecolor{lightgray}{gray}{0.75}
\usepackage{tcolorbox}

\usepackage{caption}

\usepackage{titlesec}
\titlespacing\section{0pt}{0pt plus 0pt minus 2pt}{0pt plus 0pt minus 2pt}
\titlespacing\subsection{0pt}{0pt plus 0pt minus 2pt}{0pt plus 0pt minus 2pt}
\titlespacing\subsubsection{0pt}{0pt plus 0pt minus 2pt}{0pt plus 0pt minus 2pt}

\AtBeginDocument{%
  \addtolength\abovedisplayskip{-0.3\baselineskip}%
  \addtolength\belowdisplayskip{-0.3\baselineskip}%
 \addtolength\abovedisplayshortskip{-0.3\baselineskip}%
 \addtolength\belowdisplayshortskip{-0.3\baselineskip}%
}

\title{Revisiting Kernel Attention with Correlated Gaussian Process Representation}

\author[1,3]{\href{mailto:<minhlongbui2000@gmail.com>?Subject=Your UAI 2024 paper}{Long Minh Bui}{}}
\author[1,2]{Tho Tran Huu}
\author[1]{Duy Dinh}
\author[2,*]{Tan Minh Nguyen}
\author[3,*]{Trong Nghia Hoang}
\affil[1]{%
    FPT Software AI Center
}
\affil[2]{%
    Department of Mathematics,
    National University of Singapore    
}
\affil[3]{%
    School of Electrical Engineering and Computer Science, Washington State University
}
\affil[*]{Co-last author}

\begin{document}
\maketitle
\begin{abstract}
    Transformers have increasingly become the de facto method to model sequential data with state-of-the-art performance. Due to its widespread use, being able to estimate and calibrate its modeling uncertainty is important to understand and design robust transformer models. To achieve this, previous works have used Gaussian processes (GPs) to perform uncertainty calibration for the attention units of transformers and attained notable successes. However, such approaches have to confine the transformers to the space of symmetric attention to ensure the necessary symmetric requirement of their GP's kernel specification, which reduces the representation capacity of the model. To mitigate this restriction, we propose the Correlated Gaussian Process Transformer (CGPT), a new class of transformers whose self-attention units are modeled as cross-covariance between two correlated GPs (CGPs). This allows asymmetries in attention and can enhance the representation capacity of GP-based transformers. We also derive a sparse approximation for CGP to make it scale better. Our empirical studies show that both CGP-based and sparse CGP-based transformers achieve better performance than state-of-the-art GP-based transformers on a variety of benchmark tasks. The code for our experiments is available at \url{https://github.com/MinhLong210/CGP-Transformers}.
\end{abstract}

\section{Introduction}
\label{sec:intro}

\par Transformers \citep{vaswani2017attention} have recently emerged as the preferred models in various sequence modeling tasks, including those in computer vision~\citep{al2019character,dosovitskiy2020image,ramesh2021zero,radford2021learning,9710415,liu2021video,zhao2021point,guo2021pct}, natural language processing~\citep{baevski2018adaptive,dehghani2018universal,devlin2018bert,al2019character,dai2019transformer,NEURIPS2020_1457c0d6,brown2020language}, and reinforcement learning~\citep{chen2021decision,janner2021offline}, due to its computational advantage in replacing expensive recurrence operations in recurrent neural networks~\citep{medsker2001recurrent} and long short-term memory (LSTM) networks~\citep{hochreiter1997long}  with a feed-forward attention mechanism that allows for significantly more parallelization in model training~\citep{lin2022survey,tay2022efficient}. Transformer-based pre-trained models can also be effectively adapted to new tasks with limited supervision~\citep{radford2018improving,radford2019language,devlin2018bert,yang2019xlnet,liu2019roberta}. In particular, the core component of a transformer model is the multi-head self-attention (MHSA), which captures sequential dependencies among different tokens of a single sequence by having each token represented as weighted average over (learnable) functions of other tokens whereas the (learnable) weights characterize the similarities between tokens~\citep{cho-etal-2014-learning,parikh-etal-2016-decomposable,DBLP:journals/corr/LinFSYXZB17}. Intuitively, such weights represent the amount of attention each token needs to give others to obtain its contextual representation~\citep{bahdanau2014neural,vaswani2017attention,kim2017structured}.

\par Despite its growing successes, the original transformer lacks a mechanism for uncertainty calibration which is essential to provide trustworthy predictions and enable robust, risk-averse decision making in safety-critical tasks~\citep{chen2023calibrating}. This limitation has motivated a recent line of work \citep{xue2021bayesian, tran2019bayesian} that develops uncertainty quantification techniques for transformers. Particularly, the most recent work of \citep{chen2023calibrating} in this direction has drawn a connection between the self-attention mechanism and the inference mechanism of a GP \citep{Rasmussen06}, which interestingly appears to share the same principle of building a contextual representation for an input based on its similarities to other inputs. 

This is perhaps not surprising in hindsight considering that GPs had been historically adopted to model various forms of spatio-temporal data \citep{luttinen2012efficient, hamelijnck2021spatio} and their inherent temporal dependencies based on a kernelized measure of input similarities. This also has a direct translation to the attention mechanism from the lens of kernel attention \citep{tsai2019transformer, chen2024primal}, albeit with no uncertainty quantification. Despite this interesting progress in connecting the modern literature of transformers to the classical research on GPs for uncertainty quantification, prior work~\citep{chen2023calibrating} in this direction has to compromise the representational capacity of transformers in order to make such a connection. In particular, both linear transformation functions for the query and value vectors of the self-attention in transformers would have to be tied to the same parameterization to cast attention output as the output of a GP with a valid symmetric kernel. This constraint reduces the model's performance significantly, as shown in our experiments, which is consistent with the results for original transformers reported in~\citep{tsai2019transformer}.


{\bf Our Contribution.} To mitigate such restriction, we introduce a new perspective of GP-based transformer which preserves the original modeling flexibility of self-attention, allowing the attention matrix to be asymmetrical as needed. This is achieved via characterizing the attention output not as a GP prediction but as a prediction based on cross-covariance between two CGPs, which allows kernel asymmetries while retaining the quantifiable uncertainty structure of GPs. To substantiate the above high-level idea, we have made the following technical contributions:

{\bf 1.} In section \ref{sec:method}, we derive a correspondence between the self-attention units of the multi-head self-attention (MHSA) mechanism and cross-covariance between two CGPs modeled as different affine transformations of a common latent GP where one GP distributes the function of queries while the other distributes the function of keys. 

 
{\bf 2.} In Section \ref{sec: SCGPT} we derive a sparse approximation to the above CGP-based attention unit, which removes the cubic dependence of CGPT's processing cost on the number of input tokens. We further develop a loss regularizer in terms of the log marginal likelihood of the CGPs (or sparse CGPs) which augments the conventional training loss of the transformer to balance between minimizing the uncertainty of attention output (i.e., CGP's prediction) and minimizing its induced classification performance.

{\bf 3.} In Section \ref{sec:experiments}, we empirically demonstrate that both our CGPT and its sparse approximation achieve better predictive performance than state-of-the-art kernel-attention and GP-based transformers across multiple computer vision and natural language processing benchmarks. 



\par \textbf{Notation.} For the rest of this paper, we will use the notation $\kappa(\cdot, \cdot)$ to denote a kernel function. The kernel matrices are denoted as $\mathcal{K}$. All other matrices are denoted as bold uppercase letters. We also denote a Gaussian distribution with mean $\mu$ and variance $\sigma^2$ using the notation $\mathbb{N}(\mu, \sigma^2)$. We also use subscripts to distinguish between contexts in which those kernel function and matrices are computed.




\section{Preliminaries}
\label{sec:background}
\subsection{Self-Attention}
\par Given an input sequence $\mathbf{X} = [\mathbf{x}_1, \ldots,\mathbf{x}_n]^\top \in \mathbb{R}^{n\times d}$ of $n$ $d$-dimensional vectors, the self-attention mechanism transforms it into the output sequence $\mathbf{V}^+=[\mathbf{v}^+_1,\ldots,\mathbf{v}^+_n]^\top\in \mathbb{R}^{n\times d}$ via two steps:

\textbf{Step 1.} The input $\mathbf{X}$ is linearly transformed into the query $\mathbf{Q}$, the key $\hat{\mathbf{K}}$, and the value $\mathbf{V}$ matrices,
\begin{eqnarray}
\mathbf{Q} &\triangleq& \left[\mathbf{q}_1,  \mathbf{q}_2,\ldots ,\mathbf{q}_n\right]^\top \ =\ \mathbf{X}\mathbf{W}_q^\top, \\
\mathbf{K} &\triangleq& \left[\mathbf{k}_1,  \mathbf{k}_2,\ldots, \mathbf{k}_n\right]^\top \ =\ \mathbf{X}\mathbf{W}_k^\top,  \\
\mathbf{V} &\triangleq& \left[\mathbf{v}_1,  \mathbf{v}_2,\ldots, \mathbf{v}_n\right]^\top \ =\ \mathbf{X}\mathbf{W}_v^\top,
\end{eqnarray}
where $\mathbf{W}_q, \mathbf{W}_k \in \mathbb{R}^{s\times d}$ and $\mathbf{W}_v \in \mathbb{R}^{s\times d}$ are weight matrices. Each tuple of vectors $\{\mathbf{q}_i, \mathbf{k}_i, \mathbf{v}_i\}_{i=1}^n$ comprise respectively the key, query and value vectors\footnote{For simplicity, we assume that the key, query and value vectors have the same dimension $s$.}.
    
\textbf{Step 2.} Given $\mathbf{Q}$, $\mathbf{K}$ and $\mathbf{V}$, the final output of the attention mechanism is as follow:
\begin{eqnarray}
\label{eq:attention-mat}
\mathbf{V}^+ &=& \mathrm{softmax}\left(\frac{\mathbf{Q}\mathbf{K}^\top}{\sqrt{d}}\right) \cdot \mathbf{V} \ =\ \mathbf{A}\mathbf{V},
\end{eqnarray}
where the softmax operator is applied to each row of the matrix $\mathbf{A} = \mathrm{softmax}({\mathbf{Q}}\mathbf{K}^\top/\sqrt{d})$. Here, $\mathbf{A}$ is the attention matrix. Eq.~\eqref{eq:attention-mat} details the softmax attention mechanism. 
\subsection{Multi-head Self-attention (MHSA)}
MHSA helps capture more diverse patterns in the input and increase the representation capacity of transformers. A MHSA comprises $h$ units of self-attention $\mathbf{V}^+_1, \mathbf{V}^+_2, \ldots, \mathbf{V}^+_h$ where $\mathbf{V}^+_i$ denote the output of the $i$-th self-attention unit defined above. The output of the MHSA is then computed as an affine transformation of these self-attention units,
\begin{eqnarray}
\hspace{-12mm}\mathbf{H} &\triangleq& \mathrm{MultiHead}\Big(\mathbf{V}^+_1, \mathbf{V}^+_2, \ldots, \mathbf{V}^+_h\Big) \nonumber\\
\hspace{-12mm}&=& \mathrm{Concatenate}\Big(\mathbf{V}^+_1, \mathbf{V}^+_2, \ldots, \mathbf{V}^+_h\Big) \ \mathbf{W}^{\top}_o,
\end{eqnarray}
where $\mathbf{W}_o \in \mathbb{R}^{d \times (h \cdot d)}$ is the weight matrix.

\subsection{Kernel Attention}

As the softmax attention essentially requires computing the similarity between pairs of keys and queries, \citet{tsai2019transformer} proposes kernel attention which replaces the $\mathrm{softmax}$ operator with a kernel function $\kappa(\mathbf{x}_a, \mathbf{x}_b)$. Kernel attention thus replaces Eq.~\eqref{eq:attention-mat} with
\begin{eqnarray} \label{eq:kernel attention}
\mathbf{V}^+ &=& \mathcal{K}\mathbf{V},
\end{eqnarray}
where the $(a,b)$-cell of the Gram matrix $\mathcal{K}$ takes value $\kappa(\mathbf{x}_a, \mathbf{x}_b) = \kappa_o\Big(\mathbf{x}_a\mathbf{W}_q^\top, \mathbf{x}_b\mathbf{W}_k^\top\Big)$, {\color{black} where $\kappa_o$ is a valid symmetric kernel}.

Note that even though $\kappa_o(\mathbf{z}, \mathbf{z}')$ can be selected to be symmetric, $\kappa(\mathbf{x}_a, \mathbf{x}_b)$ might not be so since
\begin{eqnarray}
\label{eq: kernel_attention-mat}
\hspace{-8mm}\kappa(\mathbf{x}_a, \mathbf{x}_b) &=&  \kappa_o\Big(\mathbf{x}_a\mathbf{W}_q^\top, \mathbf{x}_b\mathbf{W}_k^\top\Big) \nonumber\\
&\ne& \kappa_o\Big(\mathbf{x}_b\mathbf{W}_q^\top, \mathbf{x}_a\mathbf{W}_k^\top\Big) \ =\ \kappa(\mathbf{x}_b, \mathbf{x}_a).
\end{eqnarray}

Thus, to construct a valid symmetric kernel in kernel attention, the key and query matrices, $\mathbf{W}_k$ and $\mathbf{W}_q$, need to be identical, $\mathbf{W}_k = \mathbf{W}_q = \mathbf{W}$. Tying the parameters defining these matrices saves computational costs but will result in a limitation in the representation capacity of the model, as empirically shown in Section 3.2 in \citep{tsai2019transformer}, where attention with asymmetric kernels tends to outperform attention with symmetric kernels.

\subsection{Gaussian Processes}
A Gaussian process~\citep{Rasmussen06} defines a probabilistic prior over a random function $z(\mathbf{x})$ defined by mean function $m(\mathbf{x}) = 0$ and kernel function $\kappa(\mathbf{x}, \mathbf{x}')$.\footnote{For simplicity, we assume a zero mean function since we can always re-center the training outputs around $0$.} These functions induce a marginal Gaussian prior over the evaluations $\mathbf{z} = [z(\mathbf{x}_1) \ldots z(\mathbf{x}_n)]^\top$ on an arbitrary finite subset of inputs  $\{\mathbf{x}_1, \ldots, \mathbf{x}_n\}$. 

Let $\mathbf{x}_\ast$ be an unseen input whose corresponding output $z_\ast = z(\mathbf{x}_\ast)$ we wish to predict. The Gaussian prior over $[z(\mathbf{x}_1) \ldots z(\mathbf{x}_n)\ z(\mathbf{x}_\ast)]^\top$ implies the following conditional distribution:
\begin{eqnarray}
z_\ast &\triangleq& z(\mathbf{x}_\ast) \mid \mathbf{z} \nonumber\\
&\sim& \mathbb{N}\Big(\mathbf{k}_\ast^\top\mathcal{K}^{-1}\mathbf{z},\  \kappa(\mathbf{x}_\ast,\mathbf{x}_\ast) - \mathbf{k}_\ast^\top\mathcal{K}^{-1}\mathbf{k}_\ast\Big) \ ,\label{eq:1}
\end{eqnarray}
where $\mathbf{k}_\ast = [\kappa(\mathbf{x}_\ast, \mathbf{x}_1) \ldots \kappa(\mathbf{x}_\ast,\mathbf{x}_n)]^\top$ and $\mathcal{K}$ denotes the Gram matrix induced by $\kappa(\mathbf{x}, \mathbf{x}')$ on $\{\mathbf{x}_1, \ldots, \mathbf{x}_n\}$ whose value at cell $(a,b)$ is $\kappa(\mathbf{x}_a, \mathbf{x}_b)$. For noisy observation $z_i$ perturbed by Gaussian noise such that $z_i \sim \mathbb{N}(z(\mathbf{x}_i), \sigma^2)$,  Eq.~\eqref{eq:1} above can be integrated with $\mathbb{N}(\mathbf{z}, \sigma^2\mathbf{I})$ to yield:
\begin{eqnarray}
z_\ast &\triangleq& z(\mathbf{x}_\ast) \mid \mathbf{z} \nonumber\\
&\sim& \mathbb{N}\Big(\mathbf{k}_\ast^\top\mathcal{K}_{\sigma}^{-1}\mathbf{z},\  \kappa(\mathbf{x}_\ast,\mathbf{x}_\ast) - \mathbf{k}_\ast^\top \mathcal{K}_{\sigma}^{-1}\mathbf{k}_\ast\Big) \ ,\label{eq:2}
\end{eqnarray}
where $\mathcal{K}_\sigma = \mathcal{K} + \sigma^2\mathbf{I}$. Eq.~\eqref{eq:2} forms the predictive distribution of the Gaussian process (GP).

\subsection{Kernel Attention as GP Inference} \label{sec: attention as GP}
Suppose $\mathbf{X} = [\mathbf{x}_1, \ldots, \mathbf{x}_n]^\top$ is the input fed into a kernel-attention unit. Assuming that the key and query matrices are set to be identical, its output $\mathbf{V}^+ \in \mathbb{R}^{n \times s}$ is given as $\mathbf{V}^+ = \mathcal{K}\mathbf{V}=\mathcal{K}\mathbf{X}\mathbf{W}_v^\top$ where
\begin{eqnarray}
\hspace{-7.5mm}\mathcal{K}\mathbf{X}\mathbf{W}_v^\top &\triangleq& 
\mathcal{K}\Bigg(\mathcal{K} + \sigma^2\mathbf{I}\Bigg)^{-1}\mathbf{Z}. \label{eq:Z}
\end{eqnarray}
Here, we set $\mathbf{Z} = (\mathcal{K} + \sigma^2\mathbf{I})\mathbf{X}\mathbf{W}_v^\top \in \mathbb{R}^{n\times s}$ with $\mathcal{K}$ being the induced Gram matrix of $\kappa(\mathbf{x}_a, \mathbf{x}_b)$ as defined in Eq.~\eqref{eq: kernel_attention-mat} above. Thus, let $\boldsymbol{\nu}_a$ denote the $a$-th column of $\mathbf{V}^+$ and $\mathbf{z}_a$ denote the $a$-th column of $\mathbf{Z}$, we have
\begin{eqnarray}
\boldsymbol{\nu}_a &=& \mathcal{K}\Bigg(\mathcal{K} + \sigma^2\mathbf{I}\Bigg)^{-1} \mathbf{z}_a,  \label{eq:equiv}
\end{eqnarray}
or equivalently, 
$[\boldsymbol{\nu}_{a}]_r =  \mathbf{k}_r^\top(\mathcal{K} + \sigma^2\mathbf{I})^{-1} \mathbf{z}_a$ where $[\boldsymbol{\nu}_{a}]_r$ is the $r$-th component of the column vector $\boldsymbol{\nu}_a$ and $\mathbf{k}_r = [\kappa(\mathbf{x}_r, \mathbf{x}_1) \ldots \kappa(\mathbf{x}_r, \mathbf{x}_n)]^\top$. 

Comparing this to Eq.~\eqref{eq:2} earlier, it appears that the $a$-th column $\boldsymbol{\nu}_a$ of the attention output is the mean prediction on $\mathbf{x}_1, \mathbf{x}_2, \ldots, \mathbf{x}_n$ of a  modeling the dataset $\{\mathbf{x}_r, [\mathbf{z}_a]_r\}_{r=1}^n$. As such, one can assess the attention uncertainty or variance of $\boldsymbol{\nu}_a$ (i.e., the $a$-th column of $\mathbf{V}^+$),
\begin{eqnarray}
\mathbb{V}\left[\boldsymbol{\nu}_a\right] &=& \mathcal{K} \ -\  \mathcal{K}\left(\mathcal{K} + \sigma^2\mathbf{I}\right)^{-1}\mathcal{K}.
\end{eqnarray}
Overall, if the output dimension of the kernel attention unit is $s$, we can equivalently represent it using $s$ independent GPs. Furthermore, we can extend the above formalism towards multi-head self-attention with GPs by concatenating the equivalent GP inferences corresponding to each head and multiplying all with the weight matrix $\mathbf{W}_o$. 

Note that this equivalence is only possible if the kernel matrix above is symmetric, which requires $\mathbf{W}_q = \mathbf{W}_k$ as explained earlier. A more recent work by~\citep{chen2023calibrating} has also extended the above to instead align with a sparse GP inference, which similarly cast the kernel attention output in terms of the sparse GP inference. Nonetheless, like the GP attention approach, the proposed sparse GP attention will still require the use of symmetric kernel to ensure the modeling consistency of its underlying GP.

\section{Revisiting Kernel Attention}
\label{sec:method}
\par To mitigate the restriction to symmetric kernel imposed on existing GP-based transformers, we will instead explore an alternative perspective of correlated Gaussian process (CGP) modeling~\citep{aueb2013variational}. This will allow us to model the kernel attention unit in terms of the cross-covariance between two correlated Gaussian processes, which naturally permits kernel asymmetries while preserving the GP's built-in capability to calibrate uncertainty. 

This consequently inspires a principled approach to calibrate a transformer model without compromising its attention mechanism. To elaborate, Section~\ref{sec:canonical rep} and Section~\ref{sec: canonical GP} will provide background on the canonical representation of GP  and CGP modeling. Section~\ref{sec: CGP attention} will then derive a new CGP-based attention structure that can accommodate both attention asymmetries and uncertainty calibration. A sparse approximation of this CGP-based structure is further derived in Section~\ref{sec: SCGPT} for better scalability.


\subsection{Canonical Representation of GP} \label{sec:canonical rep}
Our correlated GP (CGP) modeling was inspired from a representation of GP that parameterizes its kernel in terms of an affine input scaling applied to another parameter-free, canonical GP~\citep{aueb2013variational}. We review this representation below and show how this can be extended towards correlated GP modeling.

\begin{definition}[Canonical Gaussian process (GP)] \label{def:canonical gp}
A canonical GP $z_o(\mathbf{x}) \sim \mathcal{GP}(m_o(\mathbf{x}), \kappa_o(\mathbf{x}, \mathbf{x}'))$ is a Gaussian process  specified with a zero mean function $m_o(\mathbf{x}) = 0$ and a parameter-free kernel function $\kappa_o(\mathbf{x}, \mathbf{x}')$.
\end{definition}
A canonical GP defined in Definition~\ref{def:canonical gp}, for instance, can attain a squared exponential kernel with the kernel length-scales and global scale equal to 1, 
$\kappa_o(\mathbf{x}, \mathbf{x}') = \mathrm{exp}(-0.5 \|\mathbf{x} - \mathbf{x}'\|^2)$.
Another GP $z_\lambda(\mathbf{x})$ can then be represented in terms of an affine scaling of $z_o(\mathbf{x})$,
\begin{eqnarray}
z_\lambda(\mathbf{x}) &\triangleq& \sigma_\lambda \cdot z_o\Big(\mathbf{x}\mathbf{W}^\top\Big), \label{eq:canonocal}
\end{eqnarray}
with mean $m(\mathbf{x}) = 0$ and covariance function 
\begin{eqnarray}
\kappa_\lambda(\mathbf{x}, \mathbf{x}') 
&=& \mathbb{E}\left[\sigma_\lambda^2z_o\Big(\mathbf{x}\mathbf{W}_\lambda^\top\Big)\cdot z_o\Big(\mathbf{x}'\mathbf{W}_\lambda^\top\Big)\right] \nonumber\\
&=&\sigma_\lambda^2 \kappa_o\Big(\mathbf{x}\mathbf{W}_\lambda^\top, \mathbf{x}'\mathbf{W}_\lambda^\top\Big), \label{eq:cov}
\end{eqnarray}
where the first equality follows from the definition of covariance and $z_\lambda(\mathbf{x})$ in Eq.~\eqref{eq:canonocal}, as well as the fact that $z_o(\mathbf{x})$ has zero means. The second equality holds because of the canonical kernel definition. The parameter of this kernel function is defined as a tuple $(\mathbf{W}, \sigma_\lambda)$ of parameters.

\subsection{CGP Modeling}  \label{sec: canonical GP}
Inspired by the canonical representation in Definition~\ref{def:canonical gp}, we can now characterize two GPs $z_k(\mathbf{x})$ and $z_q(\mathbf{x})$, both of which are obtained via scaling the input of $z_o(\mathbf{x})$ using the above mechanism with separate parameters $(\mathbf{W}_k, \sigma_k)$ and $(\mathbf{W}_q, \sigma_q)$. Following Eq.~\eqref{eq:cov} above, 
\begin{eqnarray} \label{eq: kernel q k}
\kappa_k(\mathbf{x}, \mathbf{x}') 
&=& \sigma_k^2 \kappa_o\Big(\mathbf{x}\mathbf{W}_k^\top, \mathbf{x}'\mathbf{W}_k^\top\Big) \nonumber\\
\kappa_q(\mathbf{x}, \mathbf{x}') 
&=& \sigma_q^2 \kappa_o\Big(\mathbf{x}\mathbf{W}_q^\top, \mathbf{x}'\mathbf{W}_q^\top\Big), \label{eq:K_k}
\end{eqnarray}
where $\kappa_o$ is a parameter-free kernel function of the canonical GP $z_o(\mathbf{x})$.
Furthermore, it can be shown that this representation also allows analytic derivation of the cross-covariance between $z_k(\mathbf{x})$ and $z_q(\mathbf{x})$ as follow, 
\begin{eqnarray} \label{eq: cross-kernel func}
\kappa_{kq}(\mathbf{x}, \mathbf{x}') &=& \sigma_k\sigma_q \ \kappa_o\Big(\mathbf{x}\mathbf{W}_k^\top, \mathbf{x}'\mathbf{W}_q^\top\Big) \nonumber\\
\kappa_{qk}(\mathbf{x}, \mathbf{x}') &=& \sigma_q\sigma_k \ \kappa_o\Big(\mathbf{x}\mathbf{W}_q^\top, \mathbf{x}'\mathbf{W}_k^\top\Big).
\end{eqnarray}
Note that, unlike the in-domain covariance functions $\kappa_k$ and $\kappa_q$, the cross-domain covariance $\kappa_{kq}$ and $\kappa_{qk}$ are not symmetric, unless we force $\mathbf{W}_k = \mathbf{W}_q = \mathbf{W}$. This relaxes the restrictive Gaussian imposition on the marginal of $(z_k(\mathbf{x}), z_q(\mathbf{x}))$ on a finite set of inputs  $\mathbf{X} = \{\mathbf{x}_1, \mathbf{x}_2, \ldots, \mathbf{x}_n\}$ while still enabling a closed-form computation of the cross-function prediction of $\mathbf{z}_q = [z_q(\mathbf{x}_1), \ldots, z_q(\mathbf{x}_n)]^\top$ given the perturbed observations $\mathbf{z}_k = [z_{k1}, z_{k2}, \ldots, z_{kn}]^\top$ with $z_{ki} \sim \mathbb{N}(z_k(\mathbf{x}_i), \sigma^2)$. This is possible because $(z_q(\mathbf{x}), z_o(\mathbf{x}))$ and $(z_k(\mathbf{x}), z_o(\mathbf{x}))$ are both Gaussian even though $(z_k(\mathbf{x}), z_q(\mathbf{x}))$ is not. This also enables a mathematical quantification of the prediction uncertainty in terms of the conditional covariance of $\mathbf{z}_q \mid \mathbf{z}_k$.

Such modeling properties are desirable because (1) the closed-form representation of the cross-function prediction will help reproduce the kernel attention form in Eq.~\eqref{eq:kernel attention} with $\mathbf{z}_q$ being the attention output and $\mathbf{z}_k$ being the input to the attention unit; and (2) the mathematically induced form of its predictive covariance can be leveraged to calibrate the uncertainty output of the attention unit. To achieve this, we will establish the closed-form prediction of $\mathbf{z}_q \mid \mathbf{z}_k$ in the rest of this section, and then establish its correspondence to (asymmetric) kernel attention in Section~\ref{sec: CGP attention}. 

To derive the closed form for $\mathbb{E}[\mathbf{z}_q \mid \mathbf{z}_k]$, note that for any set of outputs $\mathbf{z}_o = [z_o(\mathbf{x}_{o1}), z_o(\mathbf{x}_{o2}), \ldots, z_o(\mathbf{x}_{on})]$ of the canonical GP-distributed function $z_o(\mathbf{x})$ at a set of latent inputs $\mathbf{X}_o = [\mathbf{x}_{o1}, \mathbf{x}_{o2}, \ldots, \mathbf{x}_{on}]$,
\begin{eqnarray}
\hspace{-14mm}p\big(\mathbf{z}_q|\mathbf{z}_k\big) &=& \int_{\mathbf{z}_o}p(\mathbf{z}_q\mid\mathbf{z}_o) \cdot p(\mathbf{z}_o \mid \mathbf{z}_k)\ \mathrm{d}\mathbf{z}_o \ .
\end{eqnarray}
The mean of the above distribution is therefore:
\begin{eqnarray}
\hspace{-0.75mm}\mathbb{E}\Big[\mathbf{z}_q \mid \mathbf{z}_k\Big] 
\hspace{-3mm}&=&\hspace{-3mm} \int_{\mathbf{z}_o}\left(\int_{\mathbf{z}_q}\mathbf{z}_q p(\mathbf{z}_q \mid \mathbf{z}_o)\mathrm{d}\mathbf{z}_q\right) p(\mathbf{z}_o\mid\mathbf{z}_k)\ \mathrm{d}\mathbf{z}_o\ . \nonumber
\end{eqnarray}
The above can be rewritten concisely as
\begin{eqnarray}
\hspace{-13mm}\mathbb{E}\Big[\mathbf{z}_q \mid \mathbf{z}_k\Big] &=& \mathbb{E}_{\mathbf{z}_o \sim p(\mathbf{z}_o \mid \mathbf{z}_k)}\Bigg[\mathbb{E}\Big[\mathbf{z}_q \mid \mathbf{z}_o\Big] \mid \mathbf{z}_k\Bigg],\label{eq:preda}
\end{eqnarray}
where the inner expectation is over $\mathbf{z}_q \sim p(\mathbf{z}_q \mid \mathbf{z}_o)$. Now, let $\mathcal{K}_o$ denote the induced Gram matrix of $\kappa_o(\mathbf{x},\mathbf{x}')$ on the set of $n$ latent inputs $\mathbf{X}_o$. As the marginals $(z_o(\mathbf{x}), z_q(\mathbf{x}))$ and $(z_o(\mathbf{x}), z_k(\mathbf{x}))$ are both Gaussian, it follows that 
\begin{eqnarray}
\hspace{-27mm}\mathbb{E}[\mathbf{z}_q \mid \mathbf{z}_o] &=& \mathcal{K}_{qo}(\mathcal{K}_o + \sigma^2\mathbf{I})^{-1}\mathbf{z}_o \ ,\\
\hspace{-27mm}\mathbb{E}[\mathbf{z}_o \mid \mathbf{z}_k] &=& \mathcal{K}_{ok}(\mathcal{K}_k + \sigma^2\mathbf{I})^{-1}\mathbf{z}_k \ ,\label{eq:cond}
\end{eqnarray}
where $\mathcal{K}_{qo}$ and $\mathcal{K}_{ok}$ denote the cross covariance matrix between $z_q(\mathbf{x})$ and $z_o(\mathbf{x})$; and between $z_k(\mathbf{x})$ and $z_o(\mathbf{x})$ on $\mathbf{X}_o = [\mathbf{x}_{o1}, \ldots, \mathbf{x}_{on}]$, respectively. This means the entry at row $a$ and column $b$ of $\mathcal{K}_{qo}$ is $\kappa_o(\mathbf{x}_{a}\mathbf{W}_q, \mathbf{x}_{ob})$ and likewise, the entry at row $a$ and column $b$ of $\mathcal{K}_{ok}$ is $\kappa_o(\mathbf{x}_{oa}, \mathbf{x}_{b}\mathbf{W}_k)$. Eq.~\eqref{eq:cond} is the direct result of the Gaussian conditional identity. Thus, plugging Eq.~\eqref{eq:cond} into Eq.~\eqref{eq:preda} gives
\begin{eqnarray}
\hspace{-7mm}\mathbb{E}[\mathbf{z}_q \mid \mathbf{z}_k] \hspace{-2mm}&=&\hspace{-2mm} \mathbb{E}_{\mathbf{z}_o \sim p(\mathbf{z}_o\mid \mathbf{z}_k)}\Big[\mathbb{E}[\mathbf{z}_q \mid \mathbf{z}_o] \mid \mathbf{z}_k\Big] \\
\hspace{-2mm}&=&\hspace{-2mm} \mathcal{K}_{qo}(\mathcal{K}_o + \sigma^2\mathbf{I})^{-1}\mathcal{K}_{ok}(\mathcal{K}_k + \sigma^2\mathbf{I})^{-1}\mathbf{z}_k, \label{eq: exp closed form}
\end{eqnarray}
which establishes the closed form for the cross-function prediction of $\mathbf{z}_q$ conditioned on $\mathbf{z}_k$.

\subsection{Kernel Attention via CGP} \label{sec: CGP attention}
With the above cross-function GP prediction recipe, we are now ready to draw correspondence with (asymmetric) kernel attention. As usual, we have
\begin{eqnarray}
\mathbf{V}^+ &=& \mathcal{K}\mathbf{V} \ \ =\ \ \mathcal{K}\mathbf{X}\mathbf{W}_v^\top \ ,
\end{eqnarray}
where $\mathcal{K}$ corresponds to a particular choice of kernel. To draw the correspondence between this and the CGP prediction, we choose $\mathcal{K} = \mathcal{K}_{qo}(\mathcal{K}_o+\sigma^2 \mathbf{I})^{-1}\mathcal{K}_{ok}$. With this,
\begin{eqnarray}
\hspace{-14mm}\mathbf{V}^+ \hspace{-2mm}&=&\hspace{-2mm} \mathcal{K}\mathbf{V} \ =\ \mathcal{K}\mathbf{X}\mathbf{W}_v^\top \nonumber \\ 
\hspace{-7mm}\hspace{-2mm}&=&\hspace{-2mm} \mathcal{K}_{qo}(\mathcal{K}_o+\sigma^2 \mathbf{I})^{-1}\mathcal{K}_{ok}(\mathcal{K}_k+\sigma^2 \mathbf{I})^{-1}\mathbf{Z} \ ,\label{eq:cgpt_output}
\end{eqnarray}
where $\mathbf{Z} \triangleq (\mathcal{K}_k + \sigma^2\mathbf{I})\ \mathbf{X}\mathbf{W}_v^\top$. Again, let $\boldsymbol{\nu}_a$ denotes the $a$-th column of $\mathbf{V}^+$ and $\mathbf{z}_a$ denote the $a$-column of $\mathbf{Z}$,
\begin{eqnarray} 
\label{eq:cgp_attention}
\hspace{-7.5mm}\boldsymbol{\nu}_a &=& \mathcal{K}_{qo}\Big(\mathcal{K}_o+\sigma^2 \mathbf{I}\Big)^{-1}\mathcal{K}_{ok}\Big(\mathcal{K}_k+\sigma^2 \mathbf{I}\Big)^{-1}\mathbf{z}_a \ .
\end{eqnarray}
This implies the $a$-th column $\boldsymbol{\nu}_a$ of the kernel attention output in fact corresponds to the mean prediction of $z_q(\mathbf{x}_1), \ldots, z_q(\mathbf{x}_n)$ of the conditional distribution, $p(\mathbf{z}_q|\mathbf{z}_k)$ which was fitted on the dataset $\{\mathbf{x}_r, [\mathbf{z}_a]_r\}_{r=1}^n$. Here, the target fitting output $\{[\mathbf{z}_a]_r\}_{r=1}^n$ are treated as perturbed observations of $\{z_k(\mathbf{x}_r)\}_{r=1}^n$ such that $[\mathbf{z}_a]_r \sim \mathbb{N}(z_k(\mathbf{x}_r), \sigma^2)$. 

\begin{remark}[CGP-based Attention can be Asymmetric.]
Since the induced Gram matrices $\mathcal{K}_{qk}$ and $\mathcal{K}_{kq}$ do not need to be symmetric to guarantee a consistent CGP model, we are no longer constrained to set $\mathbf{W}_q = \mathbf{W}_k$ to enforce symmetric kernel. As a result, our CGP-based attention can accommodate attention asymmetries.
\end{remark}

\begin{remark}[One CGP per Attention Dimension]
Suppose the attention output is $s$-dimensional, the kernel attention unit is equivalent to a collection of $s$ CGPs modeling of $s$ datasets $\{\mathbf{x}_r, [\mathbf{z}_a]_r\}_{r=1}^n$ with $a \in [s]$. 
\end{remark}

\subsection{Correlated GP Transformer} 
\label{sec:learning kernel}
\begin{figure*}[t]
    \centering
    \includegraphics[scale=0.5]{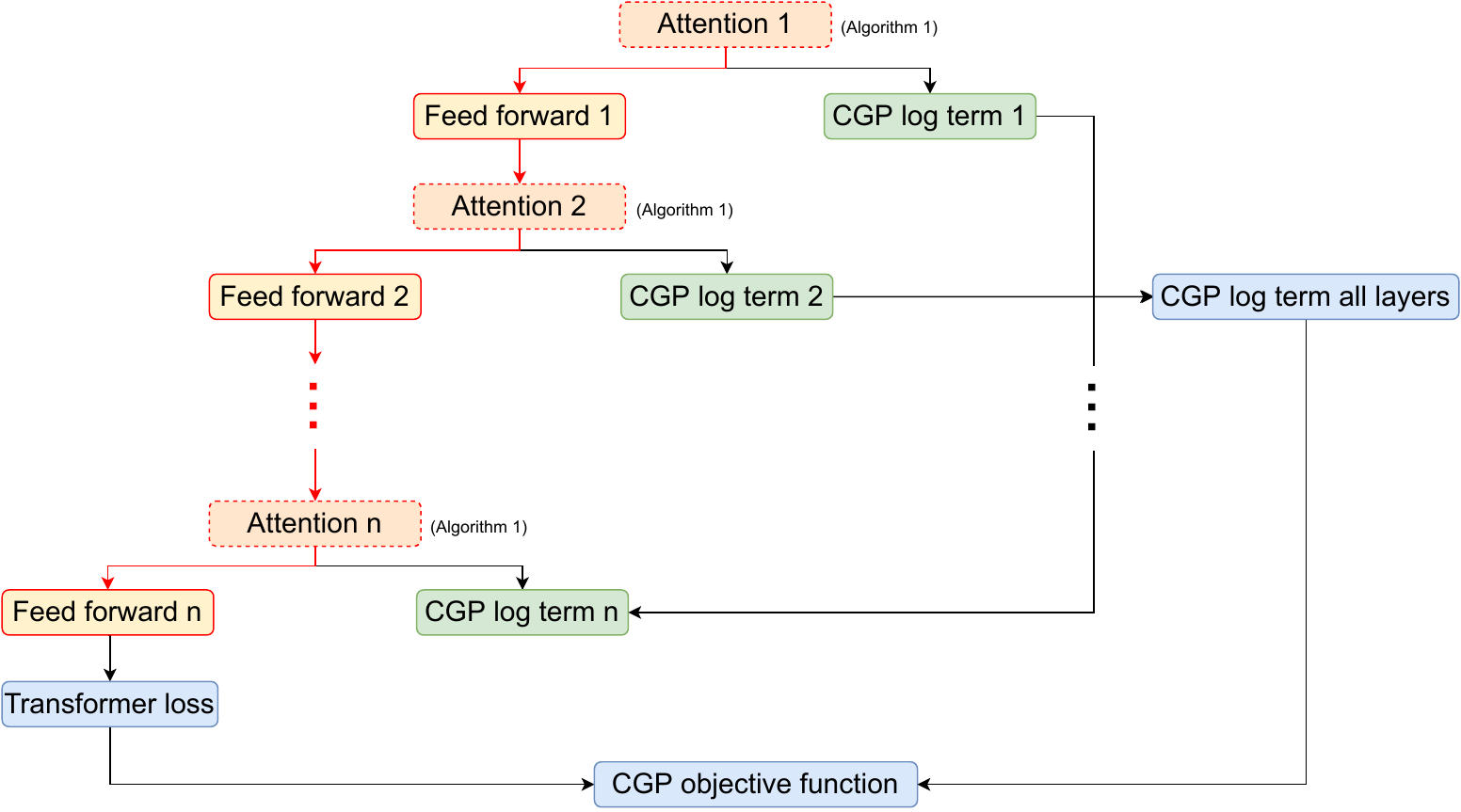}
    \caption{Diagram of the training workflow of CGPT. Each attention block forwards the CGP's prediction to the next block and caches the prediction uncertainty into a CGP regularizing term (see Algorithm~\ref{alg:cap}). Once the attention output is propagated to the last classification block, the original transformer loss is computed and augmented with the CGP regularizing term. Gradient propagation from this augmented loss will help optimize the CGP parameters to reduce prediction uncertainty while maximizing predictive performance.}
    \label{fig:diagram}
\end{figure*}

Our proposed Correlated Gaussian Process Transformer (CGPT) framework is derived via replacing the conventional attention unit with the above CGP prediction mechanism (Section~\ref{sec:cgp-attention}). Unlike a conventional transformer which optimizes for performance while neglecting uncertainty calibration for the attention output, our CGPT will optimize for both to avoid making prediction with high uncertainty while preserving overall performance. This is achieved via augmenting the original transformer loss with a regularization loss per attention block, which is expressed in terms of its prediction's uncertainty (Section~\ref{sec:CGP-regularize}). This highlights the importance of CGP's uncertainty quantification mechanism. Our overall CGPT workflow is depicted in Fig \ref{fig:diagram}.

\subsubsection{CGP-based Attention}
\label{sec:cgp-attention}
Given the above correspondence between the CGP model and kernel attention mechanism, we can replace the original kernel attention with the following CGP-based attention: 

{\bf CGP-based Attention Workflow.} At each training iteration, upon receiving $\mathbf{X} = (\mathbf{x}_1, \mathbf{x}_2, \ldots, \mathbf{x}_n)$ from the preceding neural block, we will run the following routine in Alg.~\ref{alg:cap}.
\begin{algorithm}[h!]
\caption{CGP-based Attention}\label{alg:cap}
\textbf{input:} sequence of tokens $\mathbf{X} = (\mathbf{x}_1, \mathbf{x}_2, \ldots, \mathbf{x}_n)$\\
\textbf{output:} attention output $\mathbf{V}_+$ and uncertainty $\mathcal{U}$\\\vspace{-4mm}
\begin{algorithmic}[1]
\STATE compute $\mathbf{Z} = (\mathcal{K}_k + \sigma^2\mathbf{I})\ \mathbf{X}\mathbf{W}_v^\top$ and initialize $\mathcal{U} \leftarrow 0$
\FOR{\(a \gets 1 : s\)}
\STATE build $\{\mathbf{x}_r, [\mathbf{z}_a]_r\}_{r=1}^n$ where $\mathbf{z}_a \leftarrow [\mathbf{Z}]_a$

\STATE compute $\boldsymbol{\nu}_a$ using Eq.~\eqref{eq:cgp_attention}. 

\STATE set $\mathbf{z}_k \leftarrow ([\mathbf{z}_a]_1, \ldots, [\mathbf{z}_a]_n)$

\STATE compute $\log p(\mathbf{z}_q = \boldsymbol{\nu}_a, \mathbf{z}_k = \mathbf{z}_k)$ using Eq.~\eqref{eq:CGP-loss-compute}

\STATE update $\mathcal{U} \leftarrow \mathcal{U} - \log p(\mathbf{z}_q = \boldsymbol{\nu}_a, \mathbf{z}_k = \mathbf{z}_k)$
\ENDFOR

\STATE return $\mathbf{V}_+ \leftarrow [\boldsymbol{\nu}_1, \ldots, \boldsymbol{\nu}_s]$ and $\mathcal{U}$
\end{algorithmic}
\end{algorithm}

In the above workflow, the output of each CGP-based attention block will be forwarded to the classification block that computes the logit output of the transformer, which induces its training loss. As the CGP-based attention's output is a function of the CGP's modeling parameters, the transformer loss is also a function of these parameters. Hence, the CGP can be fitted via minimizing this training loss. However, the CGP parameters learned in this manner might induced brittle prediction with high variance, especially in out-of-distribution data regime. Step $4$ of the above workflow is therefore necessary to encode a preference for parameters that induce output with low uncertainty. This is achieved via accumulating the CGP's output uncertainty -- Eq.~\eqref{eq:CGP-loss} and Eq.~\eqref{eq:CGP-loss-compute} -- per attention block into a regularizer, which is added to the main loss of the transformer prior to gradient propagation, as demonstrated in Eq.~\eqref{eq:loss}.

\subsubsection{CGP Regularization Loss}
\label{sec:CGP-regularize}
For each attention output dimension $a$, the observations $\mathbf{z}_k = [z_k(\mathbf{x}_1), z_k(\mathbf{x}_2), \ldots, z_k(\mathbf{x}_n)]$ are set to be $\mathbf{z}_a$ which is the $a$-th column of $\mathbf{Z} = (\mathcal{K}_k + \sigma^2\mathbf{I})\mathbf{X}\mathbf{W}^\top_v$. 

Following Eq.~\eqref{eq:cgp_attention}, the attention output for this dimension, $\boldsymbol{\nu}_a = \mathbb{E}[\mathbf{z}_q \mid \mathbf{z}_k]$, is the expected CGP prediction of $\mathbf{z}_q = [z_q(\mathbf{x}_1), z_q(\mathbf{x}_2),\ldots, z_q(\mathbf{x}_n)]$ given the observation $\mathbf{z}_k$. We would therefore want to maximize:
\begin{eqnarray}
\hspace{-7mm}\log p\Big(\mathbf{z}_q = \boldsymbol{\nu}_a, \mathbf{z}_k\Big) \hspace{-2mm}&=&\hspace{-2mm} \log \mathbb{E}_{\mathbf{z}_o}\Big[ p\Big(\mathbf{z}_q = \boldsymbol{\nu}_a, \mathbf{z}_k \Big | \mathbf{z}_o\Big)\Big] \label{eq:CGP-loss} 
\end{eqnarray}
because this would minimize the output uncertainty of our CGP-based attention mechanism, i.e. maximizing the fit between the input and output of the kernel attention unit. In other words, the output uncertainty of the attention output is the negation of the above log probability term. To compute it, we note that $p(\mathbf{z}_q, \mathbf{z}_k \mid \mathbf{z}_o) = p(\mathbf{z}_q \mid \mathbf{z}_o) \cdot p(\mathbf{z}_k \mid \mathbf{z}_o)$
since $\mathbf{z}_k \perp \mathbf{z}_q \mid \mathbf{z}_o$ which follows from the CGP definition. Consequently, we have
\begin{equation} \label{eq:CGP-loss-compute}
    \begin{aligned}
        &\hspace{-0.8mm}\log \mathbb{E}_{\mathbf{z}_o}\Big[p(\mathbf{z}_q, \mathbf{z}_k \mid \mathbf{z}_o)\Big] \hspace{-2.5mm}\\
        &=\log \mathbb{E}_{\mathbf{z}_o}\Big[p(\mathbf{z}_q \mid \mathbf{z}_o) \cdot p(\mathbf{z}_k \mid \mathbf{z}_o)\Big]\\
        &=\log \mathbb{E}_{\mathbf{z}_o}\Big[p(\mathbf{z}_q \mid \mathbf{z}_o)\Big] + \log \mathbb{E}_{\mathbf{z}_o}\Big[p(\mathbf{z}_k \mid \mathbf{z}_o)\Big]. 
    \end{aligned}
\end{equation}


Now, let $\mathrm{loss}(\boldsymbol{\nu}_a)$ denote the original loss of the transformer which is a function of the attention output\footnote{For simplicity, we narrate this part assuming there is a single attention block with one output dimension. Otherwise, it is straight-forward to extend the above to multiple attention blocks with multiple output dimensions by including one uncertainty term per attention block and output dimension into the final loss.} $\boldsymbol{\nu}_a$. To opt for both uncertainty minimization and performance maximization, we propose to minimize the following augmented loss $\theta_\ast = \argmin_\theta \mathfrak{L}(\theta)$ where
\begin{eqnarray}
\hspace{-10mm}\mathfrak{L}(\theta) \hspace{-2mm}&\triangleq&\hspace{-2mm} \mathrm{loss}(\boldsymbol{\nu}_a) - \alpha\cdot \log p(\mathbf{z}_q = \boldsymbol{\nu}_a, \mathbf{z}_k)  \nonumber\\
\hspace{-2mm}&=&\hspace{-2mm} \mathrm{loss}(\boldsymbol{\nu}_a) - \alpha\cdot \log \mathbb{E}_{\mathbf{z}_o}\Big[p(\mathbf{z}_q = \boldsymbol{\nu}_a, \mathbf{z}_k \mid \mathbf{z}_o)\Big],\label{eq:loss}
\end{eqnarray}
where $\alpha > 0$ is a regularization coefficient while $\theta$ represents the collection of all CGP parameters from which the attention output $\boldsymbol{\nu}_a = \mathbb{E}[\mathbf{z}_q \mid \mathbf{z}_k]$ and the CGP density $p(\mathbf{z}_q = \boldsymbol{\nu}_a, \mathbf{z}_k \mid \mathbf{z}_o)$ are computed. 

Finally, plugging Eq.~\eqref{eq:CGP-loss-compute} in Eq.~\eqref{eq:loss},
\begin{eqnarray} 
\hspace{-6mm}\mathfrak{L}(\theta)
&=& \mathrm{loss}(\boldsymbol{\nu}_a) - \alpha\cdot \Big[\log \mathbb{E}_{\mathbf{z}_o}\Big[p(\mathbf{z}_q = \boldsymbol{\nu}_a \mid \mathbf{z}_o)\Big] \nonumber\\
\hspace{-15mm}&+& \log \mathbb{E}_{\mathbf{z}_o}\Big[p(\mathbf{z}_k \mid \mathbf{z}_o)\Big] \Big] \ ,\label{eq:expanded_loss}
\end{eqnarray}
where $p(\mathbf{z}_q \mid \mathbf{z}_o)$ and $p(\mathbf{z}_k \mid \mathbf{z}_o)$ are both Gaussian whose specific forms are detailed in Appendix~\ref{app:A}. We refer to the objective function in \eqref{eq:expanded_loss} as the CGP objective and its full derivation as well as the uncertainty calibration of the induced attention output is detailed in Appendix~\ref{app:A}.

\begin{remark}
The regularization coefficient $\alpha > 0$ is a hyper-parameter balances between performance maximization and uncertainty minimization. In practice, it can be empirically selected using a validation set. 
\end{remark}

\section{Sparse Approximation} \label{sec: SCGPT}
As CGPT is developed based on the correlation structure between the two full-rank, correlated Gaussian processes, its complexity also scales cubically in the size of the number of input tokens. This is evident in Eq.~\eqref{eq:cgp_attention} which computes the CGP's predictive mean via inverting Gram matrices of size $n$ by $n$ where $n$ is the length of the input sequence. This incurs a prohibitively expensive computation cost of $\mathbb{O}(n^3)$. To mitigate this cubic dependence on the input length, we further develop a sparse approximation to CGP.
The resulting sparse approximation can thus replaced the aforementioned CGP-based attention, which gives rise to a new framework of sparse correlated Gaussian process transformer (SCGPT).

To derive this sparse approximation, we begin with the predictive mean $\mathbb{E}[\mathbf{z}_q\mid \mathbf{z}_k]$, whose explicit form is essential to draw correspondence to the output of kernel attention,
\begin{eqnarray} \label{eq: predictive step 1}
\hspace{-11mm}\mathbb{E}\Big[\mathbf{z}_q \mid \mathbf{z}_k\Big] &=& \mathbb{E}_{\mathbf{z}_o \sim p(\mathbf{z}_o \mid \mathbf{z}_k)}\Bigg[\mathbb{E}\Big[\mathbf{z}_q \mid \mathbf{z}_o\Big] \mid \mathbf{z}_k\Bigg]\ .\label{eq:s1}
\end{eqnarray}
Following the previous derivation in Section~\ref{sec: canonical GP}, we recognize that the main computational bottleneck stems from the fact that we are computing the nested expectation above with respect to the predictive distributions of two full-rank Gaussian processes, $p(\mathbf{z}_q \mid \mathbf{z}_o)$ and $p(\mathbf{z}_o \mid \mathbf{z}_k)$. Thus, to mitigate such bottleneck, we can instead adopt existing sparse approximations of Gaussian processes~\citep{Smola01,Tresp00,Tresp03,seeger2003fast,Candela05,Snelson06,Titsias09,Miguel10,Hensman13,NghiaICML15,NghiaICML16,NghiaAAAI17,NghiaAAAI19}. In this work, we use the Deterministic Training Conditional (DTC) approximation of~\citep{seeger2003fast}. 

Specifically, we first replace the exact $p(\mathbf{z}_q \mid \mathbf{z}_o)$ with its DTC approximation, which results in the following sparse approximation of $\mathbb{E}[\mathbf{z}_q \mid \mathbf{z}_o]$,
\begin{eqnarray}
\hspace{-7mm}\mathbb{E}[\mathbf{z}_q\mid\mathbf{z}_o] \hspace{-3mm}&=&\hspace{-3mm} \frac{1}{\sigma^2} \mathcal{K}_{qm}\Big (\mathcal{K}_{mm} + \frac{1}{\sigma^2} \mathcal{K}_{mo}\mathcal{K}_{om}\Big )^{-1}\hspace{-2mm}\mathcal{K}_{mo}\mathbf{z}_o, \label{eq:s2}
\end{eqnarray}
where $\mathcal{K}_{mm}$ is the Gram matrix of a set of inducing points $\{\mathbf{s}_1, \mathbf{s}_2, \ldots, \mathbf{s}_m\}$ that lives on the input space of $z_o(\mathbf{x})$ while $\mathcal{K}_{qm}$ and $\mathcal{K}_{om}$ denote the cross-covariance matrices between the $\{z_q(\mathbf{x}_i)\}_{i=1}^n$, $\{z_o(\mathbf{x}_i)\}_{i=1}^n$ with $\{\mathbf{s}_i\}_{i=1}^m$, respectively; and $\mathcal{K}_{mo}$ is the transposition of $\mathcal{K}_{om}$.

Likewise, we can do the same for $p(\mathbf{z}_o \mid \mathbf{z}_k)$, which results in the following sparse approximation for $\mathbb{E}[\mathbf{z}_o \mid \mathbf{z}_q]$,
\begin{eqnarray}
\hspace{-8.5mm}\mathbb{E}(\mathbf{z}_o\mid\mathbf{z}_k) \hspace{-2mm}&=&\hspace{-2mm} \frac{1}{\sigma^2} \mathcal{K}_{o\ell}\Big (\mathcal{K}_{\ell\ell} + \frac{1}{\sigma^2} \mathcal{K}_{\ell k}\mathcal{K}_{k\ell}\Big )^{-1}\mathcal{K}_{\ell k}\mathbf{z}_k, \label{eq:s3}
\end{eqnarray}
which is based on another set of inducing points $\{\mathbf{s}'_i\}_{i=1}^{\ell}$ that lives on the input space of $z_o(\mathbf{x})$ while $\mathcal{K}_{o\ell}$ and $\mathcal{K}_{k\ell}$ denote the cross-covariance between the $\{z_o(\mathbf{x}_i)\}_{i=1}^n$, $\{z_k(\mathbf{x}_i)\}_{i=1}^n$ with $\{\mathbf{s}'_i\}_{i=1}^{\ell}$, respectively; $\mathcal{K}_{\ell k}$ is the transposition of $\mathcal{K}_{k\ell}$ and $\mathcal{K}_{\ell\ell}$ is the Gram matrix of $\{\mathbf{s}'_i\}_{i=1}^{\ell}$.
Plugging Eq.~\eqref{eq:s2} and Eq.~\eqref{eq:s3} into Eq.~\eqref{eq:s1} leads to a closed form for the SCGP's predictive mean,
\begin{eqnarray}
\hspace{-2mm}\mathbb{E}[\mathbf{z}_q\mid\mathbf{z}_k] &=& \frac{1}{\sigma^4} \mathcal{K}_{qm}\Big (\mathcal{K}_{mm} + \frac{1}{\sigma^2} \mathcal{K}_{mo}\mathcal{K}_{om}\Big )^{-1}\mathcal{K}_{mo}\nonumber\\ &\times&\mathcal{K}_{o\ell}\Big (\mathcal{K}_{\ell\ell} + \frac{1}{\sigma^2} \mathcal{K}_{\ell k}\mathcal{K}_{k\ell}\Big )^{-1}\mathcal{K}_{\ell k}\mathbf{z}_k ,   \label{eq:s4}
\end{eqnarray}
which is a direct consequence of taking expectation of Gaussian random variables. The readers are referred to Appendix~\ref{sec: predictive mean} for a detailed step-by-step derivation. Using Eq.~\eqref{eq:s4}, we can now draw a correspondence between the predictive mean of SCGP and the output of kernel attention, 
\begin{eqnarray}
\mathbf{V}^+ &=& \mathcal{K}\mathbf{V} \ \ =\ \ \mathcal{K}\mathbf{X}\mathbf{W}_v^\top.
\end{eqnarray}
via setting $\mathbf{Z} = \mathbf{X}\mathbf{W}_v^\top$ and the kernel attention matrix $\mathcal{K}$ as
\begin{eqnarray}
\mathcal{K} &\triangleq& \frac{1}{\sigma^4} \mathcal{K}_{qm}\Big (\mathcal{K}_{mm} + \frac{1}{\sigma^2} \mathcal{K}_{mo}\mathcal{K}_{om}\Big )^{-1}\nonumber\\
&\times&\mathcal{K}_{mo}
\mathcal{K}_{o\ell}\Big (\mathcal{K}_{\ell\ell} + \frac{1}{\sigma^2} \mathcal{K}_{\ell k}\mathcal{K}_{k\ell}\Big )^{-1}\mathcal{K}_{\ell k} \ .\label{eq:s5}
\end{eqnarray}
\par 
The cubic cost is now mitigated to the number $m$ and $\ell$ of the two set of inducing inputs introduced above. Thus, to improve scalability, we can opt for small values of $m$ and $\ell$ such that $\kappa = \max(m, \ell) \ll n$. Hence, the overall complexity of computing the predictive mean of SCGP is now $\mathcal{O}(\kappa^3+n^2 \cdot \kappa)$ which is more efficient than CGPT's which is $\mathcal{O}(n^3)$.

\begin{remark}
Both sets of inducing inputs $\{\mathbf{s}_1, \mathbf{s}_2, \ldots, \mathbf{s}_m\}$ and $\{\mathbf{s}'_1, \mathbf{s}'_2, \ldots, \mathbf{s}'_{\ell}\}$ can be treated as part of the kernel parameters in our SCGP scheme and can be learned together with other kernel parameters while we optimize the corresponding augmented loss of SCGPT, whose details are deferred to Appendix~\ref{sec: SCGP objective} due to limited space.
\end{remark}

\section{Experimental Results}
\label{sec:experiments}
\begin{table*}[t!]
        \centering
        \caption{Test MCC and other uncertainty calibration metrics achieved by SGPA and our CGPT/SCGPT on the CoLA dataset under both in-distribution and out-of-distribution settings. The first rows report results on CoLA under the in-distribution setting. The last row reports results on CoLA under the OOD setting. For each metric, we report the mean value and its standard deviation obtained over multiple runs.}
        \vspace{-1em}
        \scriptsize
        \begin{tabularx}{0.665\linewidth}{|l l l l l l|} 
            \toprule
            \textbf{Dataset} & \textbf{Model} & \bf{MCC $\uparrow$} & \bf{NLL $\downarrow$} & \bf{MCE $\downarrow$} & \bf{ECE $\downarrow$}\\
            \midrule
            
            \multirow{2}{*}{CoLA} & SGPA & \textbf{28.826 $\pm$ 0.982} & 0.842 $\pm$ 0.045 & \textbf{0.713 $\pm$ 0.031} & 0.257 $\pm$ 0.011\\ 
            & CGPT (ours) & 26.471 $\pm$ 0,387 & \textbf{0.774 $\pm$ 0.010} & 0.725 $\pm$ 0.013 & \textbf{0.236 $\pm$ 0.004}\\
            & SCGPT (ours) & 27.686 $\pm$ 1.425 & 2.254 $\pm$ 0.204 & 0.724 $\pm$ 0.004 & {0.293 $\pm$ 0.003}\\
            \midrule

            \multirow{2}{*}{CoLA (OOD)} & SGPA & 22.500 $\pm$ 0.877 & 0.876 $\pm$ 0.053 & 0.740 $\pm$ 0.040 & 0.271 $\pm$ 0.0192\\ 
            & CGPT (ours) & \textbf{26.957 $\pm$ 0.748} & \textbf{0.749 $\pm$ 0.038} & {0.711 $\pm$ 0.002} & \textbf{0.230 $\pm$ 0.004}\\
            & SCGPT (ours) & {25.369 $\pm$ 0.452} & {2.243 $\pm$ 0.110} & \textbf{0.700 $\pm$ 0.003} & {0.306 $\pm$ 0.010}\\
            
            \bottomrule
        \end{tabularx}
        \label{tab:in-distribution}
        \vspace{-1.2em}
    \end{table*}
{\small
    \begin{table*}[t]
        \centering
        \caption{Test accuracy and other calibration metrics achieved by our CGPT/SCGPT models on CIFAR10-C dataset under the OOD setting. For each distortion category, we report the mean metrics over all distortion types. We observe that CGPT/SCGPT attains better accuracy and calibration metrics than SGPA across $14/16$ cases, indicating that CGPT/SCGPT is more robust than SGPA under distribution shift.}
        \vspace{-1em}
        \scriptsize
        \begin{tabularx}{0.759\linewidth}{|l l l l l l l|} 
            \toprule
            \textbf{Metric} & \textbf{Model} & \bf{Noise} & \bf{Blur} & \bf{Weather} & \bf{Digital} & \bf{Avg.}\\
            \midrule
            
            \multirow{2}{*}{Acc $\uparrow$} & SGPA & 50.803 $\pm$ 0.447 & {59.264 $\pm$ 0.915} & \textbf{64.148 $\pm$ 0.472} & \textbf{63.028 $\pm$ 0.334} & {59.722 $\pm$ 0.323}\\
            & Kernel (asym) & 53.014 $\pm$ 0.040 & \textbf{61.327 $\pm$ 1.511} & {63.426 $\pm$ 0.930} & {62.507 $\pm$ 0.847} & {60.340 $\pm$ 0.816}\\
            & Kernel (sym) & 52.675 $\pm$ 0.190 & {60.093 $\pm$ 1.846} & {62.643 $\pm$ 0.472} & {62.710 $\pm$ 0.520} & {59.884 $\pm$ 0.838}\\
            & CGPT (ours) & {55.177 $\pm$ 0.953} & 56.412 $\pm$ 1.506 & 61.515 $\pm$ 0.703 & 60.373 $\pm$ 0.123 & 58.591 $\pm$ 0.664\\
            & SCGPT (ours) & \textbf{57.701 $\pm$ 0.870} & {59.647 $\pm$ 0.925} & {63.287 $\pm$ 0.849} & {62.516 $\pm$ 0.252} & \textbf{61.746 $\pm$ 0.438}\\
            \midrule

             \multirow{2}{*}{NLL $\downarrow$} & SGPA & 3.464 $\pm$ 0.423 & 2.551 $\pm$ 0.091 & 2.137 $\pm$ 0.162 & 2.298 $\pm$ 0.045 & 2.626 $\pm$ 0.202\\ 
             & Kernel (asym) & {3.779 $\pm$ 0.604} & 2.690 $\pm$ 0.293 & {2.462 $\pm$ 0.305} & \text{2.673 $\pm$ 0.176} & {2.875 $\pm$ 0.384}\\
             & Kernel (sym) & 3.379 $\pm$ 0.448 & {2.435 $\pm$ 0.177} & 2.262 $\pm$ 0.283 & {2.389 $\pm$ 0.303} & 2.591 $\pm$ 0.331\\
            & CGPT (ours) & \textbf{1.688 $\pm$ 0.033} & \textbf{1.565 $\pm$ 0.068} & \textbf{1.352 $\pm$ 0.049} & \textbf{1.461 $\pm$ 0.027} & \textbf{1.516 $\pm$ 0.029}\\
            & SCGPT (ours) & 2.060 $\pm$ 0.064 & {1.835 $\pm$ 0.081} & {1.663 $\pm$ 0.046} & {1.796 $\pm$ 0.051} & {1.787 $\pm$ 0.017}\\
            \midrule

             \multirow{2}{*}{MCE $\downarrow$} & SGPA & 0.668$\pm$ 0.009 & 0.592 $\pm$ 0.014 & 0.576 $\pm$ 0.014 & 0.575 $\pm$ 0.001 & 0.593 $\pm$ 0.002\\ 
             & Kernel (asym) & 0.512$\pm$ 0.021 & 0.460 $\pm$ 0.016 & 0.456 $\pm$ 0.010 & 0.457 $\pm$ 0.020 & 0.470 $\pm$ 0.018\\
             & Kernel (sym) & 0.498$\pm$ 0.014 & 0.449 $\pm$ 0.011 & 0.443 $\pm$ 0.007 & 0.437 $\pm$ 0.020 & 0.456 $\pm$ 0.024\\
            & CGPT (ours) & \textbf{0.360 $\pm$ 0.011} & \textbf{0.334 $\pm$ 0.013} & \textbf{0.284 $\pm$ 0.002} & \textbf{0.314 $\pm$ 0.003} & \textbf{0.324 $\pm$ 0.002}\\
            & SCGPT (ours) & {0.443 $\pm$ 0.018} & {0.417 $\pm$ 0.016} & {0.400 $\pm$ 0.003} & {0.419 $\pm$ 0.003} & {0.421 $\pm$ 0.004}\\
            \midrule

             \multirow{2}{*}{ECE $\downarrow$} & SGPA & 0.532 $\pm$ 0.021 & 0.488 $\pm$ 0.012 & 0.469 $\pm$ 0.003 & 0.472 $\pm$ 0.010 & 0.487 $\pm$ 0.012\\ 
             & Kernel (asym) & 0.377 $\pm$ 0.015 & 0.294 $\pm$ 0.004 & 0.275 $\pm$ 0.008 & 0.280 $\pm$ 0.011 & 0.304 $\pm$ 0.009\\
             & Kernel (sym) & 0.363 $\pm$ 0.023 & 0.285 $\pm$ 0.001 & 0.266 $\pm$ 0.012 & 0.267 $\pm$ 0.012 & 0.292 $\pm$ 0.010\\
            & CGPT (ours) & \textbf{0.226 $\pm$ 0.012} & \textbf{0.202 $\pm$ 0.007} & \textbf{0.159 $\pm$ 0.004} & \textbf{0.183 $\pm$ 0.003} & \textbf{0.192 $\pm$ 0.001}\\
            & SCGPT (ours) & {0.292 $\pm$ 0.010} & {0.259 $\pm$ 0.004} & {0.234 $\pm$ 0.005} & {0.243 $\pm$ 0.004} & {0.249 $\pm$ 0.002}\\
            \bottomrule
        \end{tabularx}
        \label{tab:OOD CIFAR}
        \vspace{-0.15in}
    \end{table*}
}
{\small
\begin{table*}[t!]
            \centering
            \caption{Averaged OOD detection performance achieved by SCGPT, SGPA and kernel attention over $4$ datasets (Textures, LSUNCrop, LSUNResize and TinyImageNetCrop). For each method, the average OOD performance is reported for each detector. SCGPT outperforms the baselines in most OOD detection metrics and has the best performance on average, suggesting its advantage on OOD detection task.}
            \vspace{-1em}
            \scriptsize	
            \begin{tabularx}{\linewidth}{|X X X X X X|} 
                \toprule
                \textbf{Model} & \textbf{Detector} & \bf{AUROC $\uparrow$} & \bf{AUPR-IN $\uparrow$} & \bf{AUPR-OUT $\uparrow$} & \bf{FPR@95 $\downarrow$}\\
                \midrule

                \multirow{5}{*}{Kernel (sym)} & KLMatching & 63.80 & 60.61 & 63.70 & 87.08\\
                
                & MaxSoftmax & 69.39 & 61.00 & \textbf{75.21} & 71.68\\
                
                & Entropy & 69.82 & 62.08 & 75.35 & 71.68\\
                
                & Energy-Based & 72.83 & 62.79 & 76.85 & 65.08\\
                
                & Average & 68.21 & 61.97 & 72.03 & 73.38\\

                \midrule

                \multirow{5}{*}{Kernel (asym)} & KLMatching & 64.72 & 60.62 & 62.83 & 92.47\\
                
                & MaxSoftmax & \textbf{69.51} & \textbf{61.40} & 75.11 & \textbf{71.30}\\
                
                & Entropy & \textbf{70.01} & 62.54 & 75.39 & 70.78\\
                
                & Energy-Based & 76.15 & 66.61 & 80.96 & 58.95\\
                
                & Average & 70.10 & 62.78 & 73.32 & 73.37\\

                \midrule

                \multirow{5}{*}{SGPA} & KLMatching & 64.82 & 60.32 & 63.75 & 90.72\\
                
                & MaxSoftmax & 68.63 & 60.76 & 74.50 & 72.62\\
                
                & Entropy & 69.16 & 61.97 & 74.74 & 72.31\\
                
                & Energy-Based & \textbf{77.78} & 62.66 & \textbf{82.46} & \textbf{58.21}\\
                
                & Average & 70.09 & 61.42 & 73.86 & 73.47\\

                \midrule

                \multirow{5}{*}{SCGPT (ours)} & KLMatching & \textbf{67.31} & \textbf{62.11} & \textbf{66.78} & \textbf{86.50}\\
                
                & MaxSoftmax & 69.18 & 61.10 & 75.13 & 71.39\\
                
                & Entropy & 69.91 & \textbf{62.82} & \textbf{75.47} & 70.89\\
                
                & Energy-Based & 77.70 & \textbf{68.56} & 82.06 & 60.09\\
                
                & Average & \textbf{70.27} & \textbf{63.40} & \textbf{74.12} & \textbf{72.47} \\
                
                
                
                
                
                 
                \bottomrule
            \end{tabularx}
            \vspace{-0.2in}
            \label{tab:ood detection2}
        \end{table*}
}
\par In this section, we empirically study the advantages of CGPT and SCGPT in calibrating transformers on a variety of tasks including the COLA linguistic acceptability prediction task \citep{warstadt2019neural} and CIFAR10 classification and out-of-distribution (OOD) evaluation \citep{krizhevsky2009cifar,hendrycks2019benchmarking}. We aim to show that both of our CGPT and SCGPT can attain comparable or better calibration ability than the SGPA \citep{chen2023calibrating} and kernel attention \citep{tsai2019transformer} baselines due to the increased representation capacity, which is enabled via the use of asymmetric kernel function. Moreover, we also compare the complexity efficiency of SCGPT against the SGPA \citep{chen2023calibrating} baseline.  We adopt similar experiment settings as in \citep{chen2023calibrating}. Additional experimental results are provided in Appendix~\ref{app:C}. 

\subsection{Experiment Settings} \label{app:D}
    Following the prior work of ~\citep{chen2023calibrating}, we will conduct experiments on image classification and the linguistic acceptability prediction with the following setup:
    \par \textbf{Tasks.} We study the performance of CGPT and SCGPT on image classification using the CIFAR10 \citep{krizhevsky2009cifar} dataset and linguistic acceptability prediction using the CoLA dataset \citep{warstadt2019neural}. For the out-of-distribution (OOD) evaluations, we use the corrupted CIFAR10-C dataset \citep{hendrycks2019benchmarking} for image classification and the out-of-distribution data within the CoLA dataset for linguistic acceptability prediction. We also evaluate and compare the uncertainty calibration of our proposed models and other baselines in OOD detection tasks for image classification (see Section~\ref{sec: OOD detection}).
    
    \par \textbf{General settings for all tasks.} For SGPA and kernel attention, we use the ARD-RBF kernel \citep{Rasmussen06} for the image classification tasks
    $ \kappa(\mathbf{x}, \mathbf{x}') = \sigma_s^2 \exp({-0.5\sum_{i=1}^d (x_i-x'_i)^2/\sigma_i^2})$ , and an exponential of scaled dot product variant for the linguistic acceptability task $\kappa(\mathbf{x},\mathbf{x}')=\sigma_s^2 \exp(\sum_{i=1}^d x_ix'_i/\sigma_i^2)$ . Here, $\mathbf{x}$ 
    and $\mathbf{x}'$ are $d$-dimensional inputs, $\sigma_s^2$ denotes the output variance and $\{\sigma_i^2\}_{i=1}^d$ are the length scales. 
    For CGPT and SCGPT, we use the parameter-free squared exponential kernel function for all tasks $\kappa_o(\mathbf{x}, \mathbf{x}') = \mathrm{exp}(-0.5 \|\mathbf{x} - \mathbf{x}'\|^2)$ as the canonical representation and model the latent inputs $\mathbf{X}_o$ by linear projection of a finite set of inputs $\mathbf{X}$ for simplicity. 
    The regularization coefficient $\alpha$ in our objective function is chosen using its induced performance on a validation dataset. Our experiments are conducted on A100 40GB SMX NVIDIA GPUs.  

    \par \textbf{Baselines.} We compare the performance of our proposed models against that of SGPA~\citep{chen2023calibrating}, which leverages sparse GP to design (symmetric) attention and provide uncertainty calibration for the Transformer. Our proposed models are also compared against standard non-GP baselines with symmetric and asymmetric kernel attention \citep{tsai2019transformer}. 

\par \textbf{Architectures.} We use Vision Transformer \citep{dosovitskiy2020image} for image classification and standard transformer architecture \citep{vaswani2017attention} for linguistic acceptability prediction. We use the parameter-free squared exponential kernel for CGPT and SCGPT for both of the tasks while in SGPA, we use the ARD kernel \citep{Rasmussen06} for image classification and the exponential kernel for linguistic acceptability prediction. 
\par \textbf{Evaluation.} We study the calibration capacity of the models by evaluating the robustness of them under out-of-distribution setting in section \ref{sec:experiments}. We also compare the out-of-distribution detection capacity of our methods against other baselines in section \ref{sec: OOD detection}. We report the accuracy (Acc) for the image classification tasks and Matthew correlation coefficient (MCC) for CoLA, as well as other test calibration metrics, including negative log likelihood (NLL), expected calibration error (ECE) and maximum calibration error (MCE).

\par \textbf{CGPT and SCGPT proprietary hyperparameters.} The $\alpha$ value in our CGP objective function is linearly annealed from $0.0$ to $1.0$ during the training phase. For SCGPT, we set the inducing variable dimension $m$ to be $m=16$ in image classification tasks, which is smaller than the sequence length $n$ in order to be more memory and computationally efficient, as discussed in Section \ref{sec: SCGPT}. The value of the noise $\sigma$ in SCGPT is tuned from $0$ to $1$ and chosen to be $\sigma=0.1$ as we find that value gives the best performance for SCGPT.

    \subsubsection{Image Classification} \label{sec:details image}
    For the OOD tasks on CIFAR10-C,
    we use the corrupted datasets and the models trained on the clean datasets to evaluate the OOD performances. The CIFAR10-C dataset contains 19 types of distortions covering 4 distortion categories: Noise, Blur, Weather and Digital. For each experiment on each type of corruption, we report the mean OOD results over multiple independent runs. The corresponding standard deviations are also reported.

    \textbf{Datasets.} The original training partition of the CIFAR10 dataset is randomly split into 45,000 instances for training and 5,000 instances for validation.

    \textbf{Implementation details.}  The architecture of ViT for the CIFAR10 dataset contains 5 MHSA layers. Each layer has 4 attention heads whose the hidden dimension is set to 128.

    Both CGPT and SCGPT are trained with batch-size 100 for 600 epochs. Their loss functions are minimized using ADAM ~\citep{kingma2014adam} with an initial learning rate of 0.0005 which decays to 0.00001 linearly. We adopt the same training scheme of~\citep{chen2023calibrating} for CGPT/SCGPT: ViT with asymmetric kernel attention is trained for the first 200 epochs and its parameters are used to initialize parameters which are continued to be updated for the next 400 epochs using the CGPT's/SCGPT's loss function. For SGPA, we use the same hyper-parameter configuration as reported in~\citep{chen2023calibrating} for training.

    \textbf{Evaluation.} We choose the best model using the validation accuracy evaluated after each $10$ epochs. The reported results are averaged over multiple independent runs. Their corresponding mean and standard deviation are also reported.
    
    \subsubsection{Linguistic Acceptability} 
    For the OOD task on the COLA dataset,
    we use the provided OOD set and the model trained on the corresponding clean dataset to evaluate the robustness of model's performance. 
    
    \textbf{Datasets.} The COLA dataset contains 516 OOD samples and the original (clean) training set, which is randomly split into $7,262$ in-distribution training samples and $1,816$ in-distribution testing samples. 
    
    \textbf{Implementation details.} The architecture of Transformer for the COLA dataset has 2 MHSA layers with each layer contains 4 attention heads. The hidden dimension and embedding dimension are 256 and 128 respectively. We also use ELMO-style representation ~\citep{DBLP:conf/naacl/PetersNIGCLZ18} for the input embeddings  as in ~\citep{chen2023calibrating}.
    
    CGPT and SCGPT are trained with batch-size 32 for 50 epochs. Their loss functions are minimized using the ADAM optimizer with an initial learning rate of 0.0005 which decays to 0.00001 linearly. For SGPA, we use the same hyper-parameter configuration of  ~\citep{chen2023calibrating}. We choose the noise term to be $\sigma=0.5$ for SCGPT.
    
    \textbf{Evaluation.} The performance of the model is evaluated after $50$ training epochs. The reported performance is averaged over multiple independent runs with different random seeds. The corresponding standard deviations are also reported.

\label{sec:OOD robustness}
\subsection{Out-of-Distribution Calibration} We perform out-of-distribution (OOD) prediction under distribution perturbation on image classification (CIFAR10) and linguistic acceptability (CoLA) tasks. We use the OOD data for CoLA provided in \citep{warstadt2019neural}. For classification task, we use the corrupted CIFAR datasets (CIFAR10-C) \citep{hendrycks2019benchmarking} as OOD data, featuring images under different forms of distortion. 
    
    \par  On the OOD CoLA dataset, Table \ref{tab:in-distribution} shows that CGPT and SCGPT achieve the best performance across all metrics evaluated. SCGPT also outperforms SGPA on $2$ out of $4$ metrics including MCC on the OOD setting.
    For the vision task, we observe that SCGPT achieves the best average out-of-distribution accuracy over all forms of distortion. SCGPT also has the second-highest performance across all calibration metrics, falling slightly behind CGPT, while surpassing all other baseline models. Interestingly, SCGPT significantly outperforms SGPA while using a set of $16$ inducing inputs, which is  half the number of inducing inputs used by SGPA. As a result, SCGPT uses less GPU memory than SGPA (see Figure \ref{fig:memory_comp compare}) while achieving better results. CGPT also outperforms all baselines across all the calibration metrics and distortion types introduced in CIFAR10-C while preserving comparable accuracy as shown in Table~\ref{tab:OOD CIFAR}. These results justify that our proposed methods are more robust than SGPA and kernel attention under distribution shift in terms of both accuracy and calibration ability.

\label{sec: OOD detection}
\subsection{Out-of-Distribution Detection} 
        In this experiment, we use the CIFAR10 dataset as the in-distribution dataset for OOD detection. 
        We choose $4$ different common image datasets for the OOD detection task which includes Textures, LSUNCrop, LSUNResize and TinyImageNetCrop as our OOD datasets. For the detectors that detect outliers, we choose $4$ state-of-the-art detectors to be used in our experiments: KLMatching \citep{hendrycks2019scaling}, Maximum Softmax Probability (MaxSoftmax) \citep{hendrycks2016baseline}, Entropy Maximization (Entropy) \citep{chan2021entropy} and Energy-Based OOD Detection (EnergyBased) \citep{liu2020energy}. We use the following standard OOD detection metrics for evaluation, which includes (1) the area under the Receiver Operating Characteristic curve (AUROC), (2) the in-distribution and out-distribution area under the Precision-Recall curve (AUPR-IN and AUPR-OUT) and (3) the false positive rate when the true positive rate is equal to 95\% (FPR@95). For each method, we evaluate the OOD performance of SCGPT and SGPA measured using the above metrics on the $4$ OOD datasets and report the mean metrics for each of the $4$ detectors. All results are reported in Table \ref{tab:ood detection2}, which shows that SCGPT has the best performance in $7/16$ cases ($4$ metrics $\times$ $4$ detectors) while the rest of the baselines (including SGPA and the two variants of kernel attention) only has the best performance in no more than $4/16$ cases. Furthermore, on average, SCGPT also outperforms all other baselines across all metrics, showing the best (averaged) quality of uncertainty estimates.
        

\subsection{Reducing Oversmoothing} Oversmoothing \citep{shi2022revisiting} occurs when the output of transformers converge to a low-rank sub-space as the number of attention blocks increases, limiting the expressivity of the models. 
Fig.~\ref{fig:over_cifar10} in Appendix \ref{sec appendix: OVSMT} demonstrates that our CGPT and SCGPT help alleviate oversmoothing in transformers while the SGPA and kernel attention baselines do not. This is demonstrated via comparing the cosine similarities between the output of the attention blocks, i.e., the greater the cosine similarities, the more oversmoothing.

            
        
\label{sec: memory}
\subsection{Efficiency Analysis} 
{\small
\begin{figure}[!t]
    \centering
    \captionsetup{font=small} 
    \includegraphics[width=0.9\linewidth]{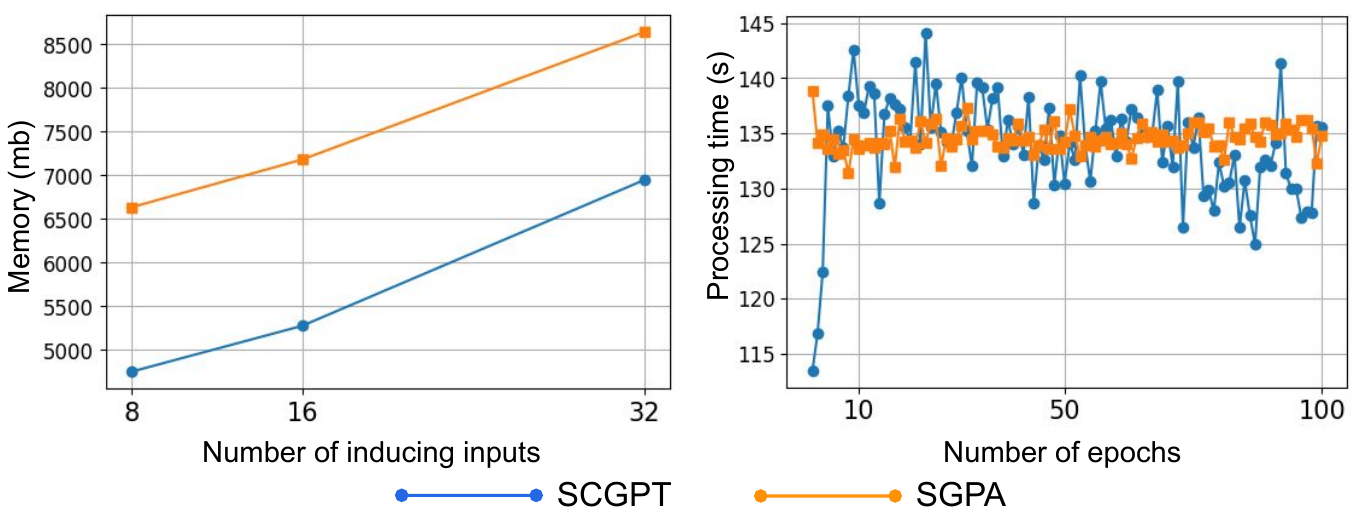}
    \vspace{-0.5em}
    \caption{Runtime per training epoch (right) and GPU memory allocated during training (left) of SCGPT and SGPA on CIFAR10. SCGPT is more efficient than SGPA in terms of GPU memory usage while having a comparable runtime per epoch to SGPA. }
    \label{fig:memory_comp compare}
    \vspace{-0.2in}
\end{figure}
}
This section compares the processing and memory costs incurred by SCGPT and SGPA during training. Since both methods utilize sparse GPs, we report their average processing time per training epoch and GPU memory usage with respect to the number of inducing inputs used in their sparse approximation. For CIFAR10, the sequence length is $64$. So, we report the processing time and memory consumption of both methods with respect to using $8$, $16$ and $32$ inducing inputs. Figure~\ref{fig:memory_comp compare} indicates that SCGPT incurs less memory cost
than SGPA while still preserving a comparable processing time to SGPA. Particularly, the GPU memory cost incurred by SCGPT is less than that of SGPA over all the above settings and the processing time per epoch of the two methods are also comparable to each other. This implies that SCGPT scales better to larger tasks than SGPA. 


\section{Related Work}
\label{sec:related_work}
Recent works have aimed to calibrate transformers using Bayesian approaches. \citet{fan2020bayesian} and ~\citet{cinquin2021pathologies} apply variational inference to the attention matrices. \citet{liu2020simple}, ~\citet{bradshaw2017adversarial} suggests fitting a GP on the output of the last attention layer. Another work utilizing GP was proposed by ~\citep{chen2023calibrating} that fits a sparse variational GP to each attention layer and propagates uncertainty across the layers. CGPT extends this research direction by fitting correlated GPs to the attention outputs.
Additionally, convolutional and recurrent neural networks, have benefited from the application of Bayesian approaches ~\citep{mukhoti2018evaluating, kendall2017uncertainties, gustafsson2020evaluating, chien2015bayesian, ritter2021sparse, tran2019bayesian}, and early efforts to employ similar methods for transformers have attained initial successes ~\citep{xue2021bayesian}. Another line of work by ~\citep{muller2021transformers} make the connection between transformers and Bayesian inference, showing that transformers can efficiently do Bayesian inference. Our proposed CGPT is complementary to those methods.

\section{Concluding Remarks}
\label{sec:conclusion}
This paper introduces the Correlated Gaussian Process Transformer (CGPT), a new framework to calibrate uncertainty for transformers.
CGPT leverages a novel CGP representation, which allows us to draw connection between the output of kernel attention often used in transformers and inference using cross-covariance between two correlated GPs defined through a latent canonical GP. With this formulation, our cross-covariance function does not have to be a symmetric kernel, which is a condition imposed on existing GP-based transformers in exchange for uncertainty calibration.
Therefore, our framework preserves the flexibility in the representation capacity of attention by allowing asymmetries in the attention unit while being able to fully utilize the uncertainty estimation ability of GPs. 
Improving the efficiency of CGPT using random features or sparse GP is an interesting future work to explore.



\begin{thebibliography}{72}
\providecommand{\natexlab}[1]{#1}
\providecommand{\url}[1]{\texttt{#1}}
\expandafter\ifx\csname urlstyle\endcsname\relax
  \providecommand{\doi}[1]{doi: #1}\else
  \providecommand{\doi}{doi: \begingroup \urlstyle{rm}\Url}\fi

\bibitem[Al-Rfou et~al.(2019)Al-Rfou, Choe, Constant, Guo, and
  Jones]{al2019character}
Rami Al-Rfou, Dokook Choe, Noah Constant, Mandy Guo, and Llion Jones.
\newblock Character-level language modeling with deeper self-attention.
\newblock In \emph{Proceedings of the AAAI Conference on Artificial
  Intelligence}, volume~33, pages 3159--3166, 2019.

\bibitem[Arnab et~al.(2021)Arnab, Dehghani, Heigold, Sun, Lučić, and
  Schmid]{9710415}
Anurag Arnab, Mostafa Dehghani, Georg Heigold, Chen Sun, Mario Lučić, and
  Cordelia Schmid.
\newblock Vivit: A video vision transformer.
\newblock In \emph{2021 IEEE/CVF International Conference on Computer Vision
  (ICCV)}, pages 6816--6826, 2021.
\newblock \doi{10.1109/ICCV48922.2021.00676}.

\bibitem[Baevski and Auli(2019)]{baevski2018adaptive}
Alexei Baevski and Michael Auli.
\newblock Adaptive input representations for neural language modeling.
\newblock In \emph{International Conference on Learning Representations}, 2019.
\newblock URL \url{https://openreview.net/forum?id=ByxZX20qFQ}.

\bibitem[Bahdanau et~al.(2014)Bahdanau, Cho, and Bengio]{bahdanau2014neural}
Dzmitry Bahdanau, Kyunghyun Cho, and Yoshua Bengio.
\newblock Neural machine translation by jointly learning to align and
  translate.
\newblock \emph{arXiv preprint arXiv:1409.0473}, 2014.

\bibitem[Bradshaw et~al.(2017)Bradshaw, Matthews, and
  Ghahramani]{bradshaw2017adversarial}
John Bradshaw, Alexander G de~G Matthews, and Zoubin Ghahramani.
\newblock Adversarial examples, uncertainty, and transfer testing robustness in
  gaussian process hybrid deep networks.
\newblock \emph{arXiv preprint arXiv:1707.02476}, 2017.

\bibitem[Brown and et~al.(2020)]{NEURIPS2020_1457c0d6}
Tom Brown and et~al.
\newblock Language models are few-shot learners.
\newblock In H.~Larochelle, M.~Ranzato, R.~Hadsell, M.~F. Balcan, and H.~Lin,
  editors, \emph{Advances in Neural Information Processing Systems}, volume~33,
  pages 1877--1901, 2020.
\newblock URL
  \url{https://proceedings.neurips.cc/paper/2020/file/1457c0d6bfcb4967418bfb8ac142f64a-Paper.pdf}.

\bibitem[Brown et~al.(2020)Brown, Mann, Ryder, Subbiah, Kaplan, Dhariwal,
  Neelakantan, Shyam, Sastry, Askell, et~al.]{brown2020language}
Tom Brown, Benjamin Mann, Nick Ryder, Melanie Subbiah, Jared~D Kaplan, Prafulla
  Dhariwal, Arvind Neelakantan, Pranav Shyam, Girish Sastry, Amanda Askell,
  et~al.
\newblock Language models are few-shot learners.
\newblock \emph{Advances in neural information processing systems},
  33:\penalty0 1877--1901, 2020.

\bibitem[Chan et~al.(2021)Chan, Rottmann, and Gottschalk]{chan2021entropy}
Robin Chan, Matthias Rottmann, and Hanno Gottschalk.
\newblock Entropy maximization and meta classification for out-of-distribution
  detection in semantic segmentation.
\newblock In \emph{Proceedings of the ieee/cvf international conference on
  computer vision}, pages 5128--5137, 2021.

\bibitem[Chen et~al.(2021)Chen, Lu, Rajeswaran, Lee, Grover, Laskin, Abbeel,
  Srinivas, and Mordatch]{chen2021decision}
Lili Chen, Kevin Lu, Aravind Rajeswaran, Kimin Lee, Aditya Grover, Misha
  Laskin, Pieter Abbeel, Aravind Srinivas, and Igor Mordatch.
\newblock Decision transformer: Reinforcement learning via sequence modeling.
\newblock \emph{Advances in neural information processing systems},
  34:\penalty0 15084--15097, 2021.

\bibitem[Chen and Li(2023)]{chen2023calibrating}
Wenlong Chen and Yingzhen Li.
\newblock Calibrating transformers via sparse gaussian processes.
\newblock In \emph{The Eleventh International Conference on Learning
  Representations}, 2023.
\newblock URL \url{https://openreview.net/forum?id=jPVAFXHlbL}.

\bibitem[Chen et~al.(2024)Chen, Tao, Tonin, and Suykens]{chen2024primal}
Yingyi Chen, Qinghua Tao, Francesco Tonin, and Johan Suykens.
\newblock Primal-attention: Self-attention through asymmetric kernel svd in
  primal representation.
\newblock \emph{Advances in Neural Information Processing Systems}, 36, 2024.

\bibitem[Chien and Ku(2015)]{chien2015bayesian}
Jen-Tzung Chien and Yuan-Chu Ku.
\newblock Bayesian recurrent neural network for language modeling.
\newblock \emph{IEEE transactions on neural networks and learning systems},
  27\penalty0 (2):\penalty0 361--374, 2015.

\bibitem[Cho et~al.(2014)Cho, van Merri{\"e}nboer, Gulcehre, Bahdanau,
  Bougares, Schwenk, and Bengio]{cho-etal-2014-learning}
Kyunghyun Cho, Bart van Merri{\"e}nboer, Caglar Gulcehre, Dzmitry Bahdanau,
  Fethi Bougares, Holger Schwenk, and Yoshua Bengio.
\newblock Learning phrase representations using {RNN} encoder{--}decoder for
  statistical machine translation.
\newblock In \emph{Proceedings of the 2014 Conference on Empirical Methods in
  Natural Language Processing ({EMNLP})}, pages 1724--1734, Doha, Qatar,
  October 2014. Association for Computational Linguistics.
\newblock \doi{10.3115/v1/D14-1179}.
\newblock URL \url{https://www.aclweb.org/anthology/D14-1179}.

\bibitem[Cinquin et~al.(2021)Cinquin, Immer, Horn, and
  Fortuin]{cinquin2021pathologies}
Tristan Cinquin, Alexander Immer, Max Horn, and Vincent Fortuin.
\newblock Pathologies in priors and inference for bayesian transformers.
\newblock \emph{arXiv preprint arXiv:2110.04020}, 2021.

\bibitem[Dai et~al.(2019)Dai, Yang, Yang, Carbonell, Le, and
  Salakhutdinov]{dai2019transformer}
Zihang Dai, Zhilin Yang, Yiming Yang, Jaime Carbonell, Quoc~V Le, and Ruslan
  Salakhutdinov.
\newblock Transformer-xl: Attentive language models beyond a fixed-length
  context.
\newblock \emph{arXiv preprint arXiv:1901.02860}, 2019.

\bibitem[Dehghani et~al.(2019)Dehghani, Gouws, Vinyals, Uszkoreit, and
  Kaiser]{dehghani2018universal}
Mostafa Dehghani, Stephan Gouws, Oriol Vinyals, Jakob Uszkoreit, and Lukasz
  Kaiser.
\newblock Universal transformers.
\newblock In \emph{International Conference on Learning Representations}, 2019.
\newblock URL \url{https://openreview.net/forum?id=HyzdRiR9Y7}.

\bibitem[Devlin et~al.(2019)Devlin, Chang, Lee, and Toutanova]{devlin2018bert}
Jacob Devlin, Ming-Wei Chang, Kenton Lee, and Kristina Toutanova.
\newblock {BERT}: Pre-training of deep bidirectional transformers for language
  understanding.
\newblock In \emph{Proceedings of the 2019 Conference of the North {A}merican
  Chapter of the Association for Computational Linguistics: Human Language
  Technologies, Volume 1 (Long and Short Papers)}, pages 4171--4186,
  Minneapolis, Minnesota, June 2019. Association for Computational Linguistics.
\newblock \doi{10.18653/v1/N19-1423}.
\newblock URL \url{https://aclanthology.org/N19-1423}.

\bibitem[Dosovitskiy et~al.(2020)Dosovitskiy, Beyer, Kolesnikov, Weissenborn,
  Zhai, Unterthiner, Dehghani, Minderer, Heigold, Gelly,
  et~al.]{dosovitskiy2020image}
Alexey Dosovitskiy, Lucas Beyer, Alexander Kolesnikov, Dirk Weissenborn,
  Xiaohua Zhai, Thomas Unterthiner, Mostafa Dehghani, Matthias Minderer, Georg
  Heigold, Sylvain Gelly, et~al.
\newblock An image is worth 16x16 words: Transformers for image recognition at
  scale.
\newblock \emph{arXiv preprint arXiv:2010.11929}, 2020.

\bibitem[Fan et~al.(2020)Fan, Zhang, Chen, and Zhou]{fan2020bayesian}
Xinjie Fan, Shujian Zhang, Bo~Chen, and Mingyuan Zhou.
\newblock Bayesian attention modules.
\newblock \emph{Advances in Neural Information Processing Systems},
  33:\penalty0 16362--16376, 2020.

\bibitem[Guo et~al.(2021)Guo, Cai, Liu, Mu, Martin, and Hu]{guo2021pct}
Meng-Hao Guo, Jun-Xiong Cai, Zheng-Ning Liu, Tai-Jiang Mu, Ralph~R Martin, and
  Shi-Min Hu.
\newblock Pct: Point cloud transformer.
\newblock \emph{Computational Visual Media}, 7\penalty0 (2):\penalty0 187--199,
  2021.

\bibitem[Gustafsson et~al.(2020)Gustafsson, Danelljan, and
  Schon]{gustafsson2020evaluating}
Fredrik~K Gustafsson, Martin Danelljan, and Thomas~B Schon.
\newblock Evaluating scalable bayesian deep learning methods for robust
  computer vision.
\newblock In \emph{Proceedings of the IEEE/CVF conference on computer vision
  and pattern recognition workshops}, pages 318--319, 2020.

\bibitem[Hamelijnck et~al.(2021)Hamelijnck, Wilkinson, Loppi, Solin, and
  Damoulas]{hamelijnck2021spatio}
Oliver Hamelijnck, William Wilkinson, Niki Loppi, Arno Solin, and Theodoros
  Damoulas.
\newblock Spatio-temporal variational gaussian processes.
\newblock \emph{Advances in Neural Information Processing Systems},
  34:\penalty0 23621--23633, 2021.

\bibitem[Hendrycks and Dietterich(2019)]{hendrycks2019benchmarking}
Dan Hendrycks and Thomas Dietterich.
\newblock Benchmarking neural network robustness to common corruptions and
  perturbations.
\newblock \emph{arXiv preprint arXiv:1903.12261}, 2019.

\bibitem[Hendrycks and Gimpel(2016)]{hendrycks2016baseline}
Dan Hendrycks and Kevin Gimpel.
\newblock A baseline for detecting misclassified and out-of-distribution
  examples in neural networks.
\newblock \emph{arXiv preprint arXiv:1610.02136}, 2016.

\bibitem[Hendrycks et~al.(2019)Hendrycks, Basart, Mazeika, Zou, Kwon,
  Mostajabi, Steinhardt, and Song]{hendrycks2019scaling}
Dan Hendrycks, Steven Basart, Mantas Mazeika, Andy Zou, Joe Kwon, Mohammadreza
  Mostajabi, Jacob Steinhardt, and Dawn Song.
\newblock Scaling out-of-distribution detection for real-world settings.
\newblock \emph{arXiv preprint arXiv:1911.11132}, 2019.

\bibitem[Hensman et~al.(2013)Hensman, Fusi, and Lawrence]{Hensman13}
J.~Hensman, N.~Fusi, and N.~D. Lawrence.
\newblock Gaussian processes for big data.
\newblock In \emph{Proc. UAI}, pages 282--290, 2013.

\bibitem[Hoang et~al.(2017)Hoang, Hoang, and Low]{NghiaAAAI17}
Q.~M. Hoang, T.~N. Hoang, and K.~H. Low.
\newblock A generalized stochastic variational {B}ayesian hyperparameter
  learning framework for sparse spectrum {G}aussian process regression.
\newblock In \emph{Proc. {AAAI}}, pages 2007--2014, 2017.

\bibitem[Hoang et~al.(2015)Hoang, Hoang, and Low]{NghiaICML15}
T.~N. Hoang, Q.~M. Hoang, and K.~H. Low.
\newblock A unifying framework of anytime sparse {Gaussian} process regression
  models with stochastic variational inference for big data.
\newblock In \emph{Proc. {ICML}}, pages 569--578, 2015.

\bibitem[Hoang et~al.(2016)Hoang, Hoang, and Low]{NghiaICML16}
T.~N. Hoang, Q.~M. Hoang, and K.~H. Low.
\newblock A distributed variational inference framework for unifying parallel
  sparse {G}aussian process regression models.
\newblock In \emph{Proc. {ICML}}, pages 382--391, 2016.

\bibitem[Hoang et~al.(2019)Hoang, Hoang, Low, and How]{NghiaAAAI19}
T.~N. Hoang, Q.~M. Hoang, K.~H. Low, and J.~P. How.
\newblock Collective online learning of {G}aussian processes in massive
  multi-agent systems.
\newblock In \emph{Proc. {AAAI}}, 2019.

\bibitem[Hochreiter and Schmidhuber(1997)]{hochreiter1997long}
Sepp Hochreiter and J{\"u}rgen Schmidhuber.
\newblock Long short-term memory.
\newblock \emph{Neural computation}, 9\penalty0 (8):\penalty0 1735--1780, 1997.

\bibitem[Janner et~al.(2021)Janner, Li, and Levine]{janner2021offline}
Michael Janner, Qiyang Li, and Sergey Levine.
\newblock Offline reinforcement learning as one big sequence modeling problem.
\newblock \emph{Advances in neural information processing systems},
  34:\penalty0 1273--1286, 2021.

\bibitem[Kendall and Gal(2017)]{kendall2017uncertainties}
Alex Kendall and Yarin Gal.
\newblock What uncertainties do we need in bayesian deep learning for computer
  vision?
\newblock \emph{Advances in neural information processing systems}, 30, 2017.

\bibitem[Kim et~al.(2017)Kim, Denton, Hoang, and Rush]{kim2017structured}
Yoon Kim, Carl Denton, Luong Hoang, and Alexander~M Rush.
\newblock Structured attention networks.
\newblock \emph{arXiv preprint arXiv:1702.00887}, 2017.

\bibitem[Kingma and Ba(2014)]{kingma2014adam}
Diederik~P Kingma and Jimmy Ba.
\newblock Adam: A method for stochastic optimization.
\newblock \emph{arXiv preprint arXiv:1412.6980}, 2014.

\bibitem[Krizhevsky et~al.(2009)Krizhevsky, Nair, and
  Hinton]{krizhevsky2009cifar}
Alex Krizhevsky, Vinod Nair, and Geoffrey Hinton.
\newblock Cifar-10 and cifar-100 datasets.
\newblock \emph{URl: https://www. cs. toronto. edu/kriz/cifar. html},
  6\penalty0 (1):\penalty0 1, 2009.

\bibitem[{L\'{a}zaro}-Gredilla et~al.(2010){L\'{a}zaro}-Gredilla,
  {Qui\~{n}onero}-Candela, Rasmussen, and Figueiras-Vidal]{Miguel10}
M.~{L\'{a}zaro}-Gredilla, J.~{Qui\~{n}onero}-Candela, C.~E. Rasmussen, and
  A.~R. Figueiras-Vidal.
\newblock Sparse spectrum {G}aussian process regression.
\newblock \emph{Journal of Machine Learning Research}, pages 1865--1881, 2010.

\bibitem[Lin et~al.(2022)Lin, Wang, Liu, and Qiu]{lin2022survey}
Tianyang Lin, Yuxin Wang, Xiangyang Liu, and Xipeng Qiu.
\newblock A survey of transformers.
\newblock \emph{AI Open}, 2022.

\bibitem[Lin et~al.(2017)Lin, Feng, dos Santos, Yu, Xiang, Zhou, and
  Bengio]{DBLP:journals/corr/LinFSYXZB17}
Zhouhan Lin, Minwei Feng, C{\'{\i}}cero~Nogueira dos Santos, Mo~Yu, Bing Xiang,
  Bowen Zhou, and Yoshua Bengio.
\newblock A structured self-attentive sentence embedding.
\newblock \emph{CoRR}, abs/1703.03130, 2017.
\newblock URL \url{http://arxiv.org/abs/1703.03130}.

\bibitem[Liu et~al.(2020{\natexlab{a}})Liu, Lin, Padhy, Tran, Bedrax~Weiss, and
  Lakshminarayanan]{liu2020simple}
Jeremiah Liu, Zi~Lin, Shreyas Padhy, Dustin Tran, Tania Bedrax~Weiss, and
  Balaji Lakshminarayanan.
\newblock Simple and principled uncertainty estimation with deterministic deep
  learning via distance awareness.
\newblock \emph{Advances in Neural Information Processing Systems},
  33:\penalty0 7498--7512, 2020{\natexlab{a}}.

\bibitem[Liu et~al.(2020{\natexlab{b}})Liu, Wang, Owens, and Li]{liu2020energy}
Weitang Liu, Xiaoyun Wang, John Owens, and Yixuan Li.
\newblock Energy-based out-of-distribution detection.
\newblock \emph{Advances in neural information processing systems},
  33:\penalty0 21464--21475, 2020{\natexlab{b}}.

\bibitem[Liu et~al.(2019)Liu, Ott, Goyal, Du, Joshi, Chen, Levy, Lewis,
  Zettlemoyer, and Stoyanov]{liu2019roberta}
Yinhan Liu, Myle Ott, Naman Goyal, Jingfei Du, Mandar Joshi, Danqi Chen, Omer
  Levy, Mike Lewis, Luke Zettlemoyer, and Veselin Stoyanov.
\newblock Roberta: A robustly optimized bert pretraining approach.
\newblock \emph{arXiv preprint arXiv:1907.11692}, 2019.

\bibitem[Liu et~al.(2022)Liu, Ning, Cao, Wei, Zhang, Lin, and Hu]{liu2021video}
Ze~Liu, Jia Ning, Yue Cao, Yixuan Wei, Zheng Zhang, Stephen Lin, and Han Hu.
\newblock Video swin transformer.
\newblock In \emph{I{EEE} {C}onference on {C}omputer {V}ision and {P}attern
  {R}ecognition (CVPR)}, 2022.

\bibitem[Luttinen and Ilin(2012)]{luttinen2012efficient}
Jaakko Luttinen and Alexander Ilin.
\newblock Efficient gaussian process inference for short-scale spatio-temporal
  modeling.
\newblock In \emph{Artificial Intelligence and Statistics}, pages 741--750.
  PMLR, 2012.

\bibitem[Medsker and Jain(2001)]{medsker2001recurrent}
Larry~R Medsker and LC~Jain.
\newblock Recurrent neural networks.
\newblock \emph{Design and Applications}, 5:\penalty0 64--67, 2001.

\bibitem[Mukhoti and Gal(2018)]{mukhoti2018evaluating}
Jishnu Mukhoti and Yarin Gal.
\newblock Evaluating bayesian deep learning methods for semantic segmentation.
\newblock \emph{arXiv preprint arXiv:1811.12709}, 2018.

\bibitem[M{\"u}ller et~al.(2021)M{\"u}ller, Hollmann, Arango, Grabocka, and
  Hutter]{muller2021transformers}
Samuel M{\"u}ller, Noah Hollmann, Sebastian~Pineda Arango, Josif Grabocka, and
  Frank Hutter.
\newblock Transformers can do bayesian inference.
\newblock \emph{arXiv preprint arXiv:2112.10510}, 2021.

\bibitem[Parikh et~al.(2016)Parikh, T{\"a}ckstr{\"o}m, Das, and
  Uszkoreit]{parikh-etal-2016-decomposable}
Ankur Parikh, Oscar T{\"a}ckstr{\"o}m, Dipanjan Das, and Jakob Uszkoreit.
\newblock A decomposable attention model for natural language inference.
\newblock In \emph{Proceedings of the 2016 Conference on Empirical Methods in
  Natural Language Processing}, pages 2249--2255, Austin, Texas, November 2016.
  Association for Computational Linguistics.
\newblock \doi{10.18653/v1/D16-1244}.
\newblock URL \url{https://www.aclweb.org/anthology/D16-1244}.

\bibitem[Peters et~al.(2018)Peters, Neumann, Iyyer, Gardner, Clark, Lee, and
  Zettlemoyer]{DBLP:conf/naacl/PetersNIGCLZ18}
Matthew~E. Peters, Mark Neumann, Mohit Iyyer, Matt Gardner, Christopher Clark,
  Kenton Lee, and Luke Zettlemoyer.
\newblock Deep contextualized word representations.
\newblock In Marilyn~A. Walker, Heng Ji, and Amanda Stent, editors,
  \emph{Proceedings of the 2018 Conference of the North American Chapter of the
  Association for Computational Linguistics: Human Language Technologies,
  {NAACL-HLT} 2018, New Orleans, Louisiana, USA, June 1-6, 2018, Volume 1 (Long
  Papers)}, pages 2227--2237. Association for Computational Linguistics, 2018.
\newblock \doi{10.18653/v1/n18-1202}.
\newblock URL \url{https://doi.org/10.18653/v1/n18-1202}.

\bibitem[{Qui\~{n}onero}-Candela and Rasmussen(2005)]{Candela05}
J.~{Qui\~{n}onero}-Candela and C.~E. Rasmussen.
\newblock A unifying view of sparse approximate {Gaussian} process regression.
\newblock \emph{Journal of Machine Learning Research}, 6:\penalty0 1939--1959,
  2005.

\bibitem[Radford et~al.(2018)Radford, Narasimhan, Salimans, and
  Sutskever]{radford2018improving}
Alec Radford, Karthik Narasimhan, Tim Salimans, and Ilya Sutskever.
\newblock Improving language understanding by generative pre-training.
\newblock \emph{OpenAI report}, 2018.

\bibitem[Radford et~al.(2019)Radford, Wu, Child, Luan, Amodei, and
  Sutskever]{radford2019language}
Alec Radford, Jeffrey Wu, Rewon Child, David Luan, Dario Amodei, and Ilya
  Sutskever.
\newblock Language models are unsupervised multitask learners.
\newblock \emph{OpenAI blog}, 1\penalty0 (8):\penalty0 9, 2019.

\bibitem[Radford et~al.(2021)Radford, Kim, Hallacy, Ramesh, Goh, Agarwal,
  Sastry, Askell, Mishkin, Clark, et~al.]{radford2021learning}
Alec Radford, Jong~Wook Kim, Chris Hallacy, Aditya Ramesh, Gabriel Goh,
  Sandhini Agarwal, Girish Sastry, Amanda Askell, Pamela Mishkin, Jack Clark,
  et~al.
\newblock Learning transferable visual models from natural language
  supervision.
\newblock In \emph{International Conference on Machine Learning}, pages
  8748--8763. PMLR, 2021.

\bibitem[Ramesh et~al.(2021)Ramesh, Pavlov, Goh, Gray, Voss, Radford, Chen, and
  Sutskever]{ramesh2021zero}
Aditya Ramesh, Mikhail Pavlov, Gabriel Goh, Scott Gray, Chelsea Voss, Alec
  Radford, Mark Chen, and Ilya Sutskever.
\newblock Zero-shot text-to-image generation.
\newblock In \emph{International Conference on Machine Learning}, pages
  8821--8831. PMLR, 2021.

\bibitem[Rasmussen and Williams(2006)]{Rasmussen06}
C.~E. Rasmussen and C.~K.~I. Williams.
\newblock \emph{Gaussian Processes for Machine Learning}.
\newblock MIT Press, 2006.

\bibitem[Ritter et~al.(2021)Ritter, Kukla, Zhang, and Li]{ritter2021sparse}
Hippolyt Ritter, Martin Kukla, Cheng Zhang, and Yingzhen Li.
\newblock Sparse uncertainty representation in deep learning with inducing
  weights.
\newblock \emph{Advances in Neural Information Processing Systems},
  34:\penalty0 6515--6528, 2021.

\bibitem[Schwaighofer and Tresp(2003)]{Tresp03}
A.~Schwaighofer and V.~Tresp.
\newblock Transductive and inductive methods for approximate {Gaussian} process
  regression.
\newblock In \emph{Proc. NIPS}, pages 953--960, 2003.

\bibitem[Seeger et~al.(2003)Seeger, Williams, and Lawrence]{seeger2003fast}
Matthias~W Seeger, Christopher~KI Williams, and Neil~D Lawrence.
\newblock Fast forward selection to speed up sparse gaussian process
  regression.
\newblock In \emph{International Workshop on Artificial Intelligence and
  Statistics}, pages 254--261. PMLR, 2003.

\bibitem[Shi et~al.(2022)Shi, Gao, Xu, Liang, Li, Kong, Lee, and
  Kwok]{shi2022revisiting}
Han Shi, Jiahui Gao, Hang Xu, Xiaodan Liang, Zhenguo Li, Lingpeng Kong, Stephen
  Lee, and James~T Kwok.
\newblock Revisiting over-smoothing in bert from the perspective of graph.
\newblock \emph{arXiv preprint arXiv:2202.08625}, 2022.

\bibitem[Smola and Bartlett(2001)]{Smola01}
A.~J. Smola and P.~L. Bartlett.
\newblock Sparse greedy {G}aussian process regression.
\newblock In \emph{Proc. NIPS}, pages 619--625, 2001.

\bibitem[Snelson and Gharahmani(2005)]{Snelson06}
E.~Snelson and Z.~Gharahmani.
\newblock Sparse {G}aussian processes using pseudo-inputs.
\newblock In \emph{Proc. NIPS}, pages 1259--1266, 2005.

\bibitem[Tay et~al.(2022)Tay, Dehghani, Bahri, and Metzler]{tay2022efficient}
Yi~Tay, Mostafa Dehghani, Dara Bahri, and Donald Metzler.
\newblock Efficient transformers: A survey.
\newblock \emph{ACM Computing Surveys}, 55\penalty0 (6):\penalty0 1--28, 2022.

\bibitem[Titsias(2009)]{Titsias09}
M.~K. Titsias.
\newblock Variational learning of inducing variables in sparse {G}aussian
  processes.
\newblock In \emph{Proc. {AISTATS}}, 2009.

\bibitem[Titsias et~al.(2013)Titsias, L{\'a}zaro-Gredilla,
  et~al.]{aueb2013variational}
Michalis Titsias, Miguel L{\'a}zaro-Gredilla, et~al.
\newblock Variational inference for mahalanobis distance metrics in gaussian
  process regression.
\newblock \emph{Advances in Neural Information Processing Systems}, 26, 2013.

\bibitem[Tran et~al.(2019)Tran, Dusenberry, Van Der~Wilk, and
  Hafner]{tran2019bayesian}
Dustin Tran, Mike Dusenberry, Mark Van Der~Wilk, and Danijar Hafner.
\newblock Bayesian layers: A module for neural network uncertainty.
\newblock \emph{Advances in neural information processing systems}, 32, 2019.

\bibitem[Tresp(2000)]{Tresp00}
V.~Tresp.
\newblock A {B}ayesian committee machine.
\newblock \emph{Neural Computation}, 12:\penalty0 2719--2741, 2000.

\bibitem[Tsai et~al.(2019)Tsai, Bai, Yamada, Morency, and
  Salakhutdinov]{tsai2019transformer}
Yao-Hung~Hubert Tsai, Shaojie Bai, Makoto Yamada, Louis-Philippe Morency, and
  Ruslan Salakhutdinov.
\newblock Transformer dissection: A unified understanding of transformer's
  attention via the lens of kernel.
\newblock \emph{arXiv preprint arXiv:1908.11775}, 2019.

\bibitem[Vaswani et~al.(2017)Vaswani, Shazeer, Parmar, Uszkoreit, Jones, Gomez,
  Kaiser, and Polosukhin]{vaswani2017attention}
Ashish Vaswani, Noam Shazeer, Niki Parmar, Jakob Uszkoreit, Llion Jones,
  Aidan~N Gomez, Lukasz Kaiser, and Illia Polosukhin.
\newblock Attention is all you need.
\newblock In \emph{Advances in neural information processing systems}, pages
  5998--6008, 2017.

\bibitem[Warstadt et~al.(2019)Warstadt, Singh, and Bowman]{warstadt2019neural}
Alex Warstadt, Amanpreet Singh, and Samuel~R Bowman.
\newblock Neural network acceptability judgments.
\newblock \emph{Transactions of the Association for Computational Linguistics},
  7:\penalty0 625--641, 2019.

\bibitem[Xue et~al.(2021)Xue, Yu, Xu, Liu, Hu, Ye, Geng, Liu, and
  Meng]{xue2021bayesian}
Boyang Xue, Jianwei Yu, Junhao Xu, Shansong Liu, Shoukang Hu, Zi~Ye, Mengzhe
  Geng, Xunying Liu, and Helen Meng.
\newblock Bayesian transformer language models for speech recognition.
\newblock In \emph{ICASSP 2021-2021 IEEE International Conference on Acoustics,
  Speech and Signal Processing (ICASSP)}, pages 7378--7382. IEEE, 2021.

\bibitem[Yang et~al.(2019)Yang, Dai, Yang, Carbonell, Salakhutdinov, and
  Le]{yang2019xlnet}
Zhilin Yang, Zihang Dai, Yiming Yang, Jaime Carbonell, Ruslan Salakhutdinov,
  and Quoc~V Le.
\newblock Xlnet: Generalized autoregressive pretraining for language
  understanding.
\newblock \emph{arXiv preprint arXiv:1906.08237}, 2019.

\bibitem[Zhao et~al.(2021)Zhao, Jiang, Jia, Torr, and Koltun]{zhao2021point}
Hengshuang Zhao, Li~Jiang, Jiaya Jia, Philip~HS Torr, and Vladlen Koltun.
\newblock Point transformer.
\newblock In \emph{Proceedings of the IEEE/CVF International Conference on
  Computer Vision}, pages 16259--16268, 2021.

\end{thebibliography}

\newpage
\appendix
\onecolumn
\begin{center}
{\Large \bf Supplementary Materials for}
\end{center}
\vspace{-0.4cm}
\begin{center}
\large \bf \textit{Revisiting Kernel Attention with Correlated Gaussian Process Representation}
\end{center}

\DoToC

\newpage

\section{Derivation of CGP Objective Function in Eq.~\ref{eq:expanded_loss}}
\label{app:A}

This section derives a more specific expression for our objective function in Eq.~\eqref{eq:expanded_loss}, which is quoted below
\begin{eqnarray} \label{eq:maximize}
\min_\theta \Big\{ \mathfrak{L}(\theta) &\triangleq& \mathrm{loss}(\boldsymbol{\nu}_a) - \alpha \cdot \Big(\log \mathbb{E}_{\mathbf{z}_o}\big[p(\mathbf{z}_q = \boldsymbol{\nu}_a \mid \mathbf{z}_o)\big] \ +\ \log \mathbb{E}_{\mathbf{z}_o}\big[p(\mathbf{z}_k \mid \mathbf{z}_o)\big]\Big) \Big\}  \ .
\end{eqnarray}
where $\boldsymbol{\nu}_a$ and $\mathbf{z}_a$ is defined previously in Eq.~\eqref{eq:cgp_attention}. \\\\
To sidestep the intractability of $\log \mathbb{E}[p(\mathbf{z}_q = \boldsymbol{\nu}_a \mid \mathbf{z}_o)]$ and $\log \mathbb{E}[p(\mathbf{z}_k \mid \mathbf{z}_0)]$, we will instead optimize their lower bounds as follow. First, recall that
\begin{eqnarray}
p\Big(\mathbf{z}_q = \boldsymbol{\nu}_a \mid \mathbf{z}_o\Big)  
&=&  \mathcal{N}\Big(\boldsymbol{\nu}_a\ ;\  \mathcal{K}_{qo}\left(\mathcal{K}_o + \sigma^2 \mathbf{I}\right)^{-1}\mathbf{z}_o, \ \mathcal{K}_q-\mathcal{K}_{qo}\left(\mathcal{K}_o+\sigma^2 \mathbf{I}\right)^{-1}\mathcal{K}_{oq}\Big),
\end{eqnarray}
which follows from the CGP definition. Next, let $\boldsymbol{\Sigma}_{q} \triangleq \mathcal{K}_q-\mathcal{K}_{qo}\left(\mathcal{K}_o+\sigma^2 \mathbf{I}\right)^{-1}\mathcal{K}_{oq}$ and $\mathbf{m}_q \triangleq \mathcal{K}_{qo}\left(\mathcal{K}_o + \sigma^2 \mathbf{I}\right)^{-1}\mathbf{z}_o$. Using Jensen inequality, 
\begin{eqnarray}
\hspace{-9mm}\log \mathbb{E}\big[p(\mathbf{z}_q = \boldsymbol{\nu}_a\mid\mathbf{z}_o)\big] &\geq& \mathbb{E}\big[\log p(\mathbf{z}_q = \boldsymbol{\nu}_a\mid\mathbf{z}_o)\big] \\
\hspace{-2mm}&=&\hspace{-2mm} 0.5 \cdot \mathbb{E}_{\mathbf{z}_o}\Big[-(\boldsymbol{\nu}_a-\mathbf{m}_q)^\top \boldsymbol{\Sigma}_{q}^{-1}(\boldsymbol{\nu}_a-\mathbf{m}_{q}) - \log \mathrm{det}\big(\boldsymbol{\Sigma}_{q}\big)-n\log 2\pi\Big]\\
\hspace{-2mm}&=&\hspace{-2mm} -0.5\cdot\int_{\mathbf{z}_o} p(\mathbf{z}_0)\Big[(\boldsymbol{\nu}_a - \mathbf{m}_{q})^\top \boldsymbol{\Sigma}_{q}^{-1}(\boldsymbol{\nu}_a-\mathbf{m}_{q}) + \log \mathrm{det}(\boldsymbol{\Sigma}_{q})+n\log 2\pi\Big]\mathrm{d}\mathbf{z}_0\\
\hspace{-2mm}&=&\hspace{-2mm} -0.5\cdot\int_{\mathbf{z}_0} p(\mathbf{z}_0)\Big[(\boldsymbol{\nu}_a-\mathbf{m}_{q})^\top \boldsymbol{\Sigma}_{q}^{-1}(\boldsymbol{\nu}_a-\mathbf{m}_{q})\Big]\mathrm{d}\mathbf{z}_o -0.5 \cdot \Big(\log \mathrm{det}(\boldsymbol{\Sigma}_{q}) - \log 2\pi\Big) \ .
\end{eqnarray}  
Finally, the integral in the above lower-bound can be approximated arbitrarily closely via an empirical average based on a sufficiently large number of samples $\mathbf{z}_o^i \sim p(\mathbf{z}_o) =\mathcal{N}(\mathbf{0}, \mathcal{K}_o)$. Thus, approximately, we have the following lower-bound
\begin{eqnarray}
\hspace{-14mm}\log \mathbb{E}\Big[p(\mathbf{z}_q = \boldsymbol{\nu}_a\mid\mathbf{z}_o)\Big] \hspace{-2mm}&\geq&\hspace{-2mm} -0.5\cdot \frac{1}{n}\sum_{i=1}^n \Big[\left(\boldsymbol{\nu}_a-\mathbf{m}_{q}^i\right)^\top \boldsymbol{\Sigma}_{q}^{-1}\left(\boldsymbol{\nu}_a-\mathbf{m}_{q}^i\right)\Big]  - 0.5\cdot \log \mathrm{det}(\boldsymbol{\Sigma}_{q}) - 0.5\cdot n\log 2\pi,
\end{eqnarray}
where $\mathbf{m}_q^i \triangleq \mathcal{K}_{qo}\left(\mathcal{K}_o + \sigma^2 \mathbf{I}\right)^{-1}\mathbf{z}_o^i$.
Likewise, we can also lower bound $\log \mathbb{E}[p(\mathbf{z}_k\mid\mathbf{z}_o)]$:
\begin{eqnarray}
\hspace{-22mm}\log \mathbb{E}\Big[p(\mathbf{z}_k\mid\mathbf{z}_o)\Big] \hspace{-2mm}&\geq&\hspace{-2mm} -0.5\cdot \frac{1}{n}\sum_{i=1}^n \Big[\left(\mathbf{z}_k-\mathbf{m}_k^i\right)^\top \boldsymbol{\Sigma}_{k}^{-1}\left(\mathbf{z}_k-\mathbf{m}_k^i\right)\Big]  - 0.5 \cdot \log \mathrm{det}(\boldsymbol{\Sigma}_{k}) - 0.5 \cdot n\log 2\pi,
\end{eqnarray}
where $\mathbf{m}_k^i \triangleq \mathcal{K}_{ko}\left(\mathcal{K}_o + \sigma^2 \mathbf{I}\right)^{-1}\mathbf{z}_o^i$ and $\boldsymbol{\Sigma}_{k} \triangleq \mathcal{K}_k-\mathcal{K}_{ko}\left(\mathcal{K}_o+\sigma^2 \mathbf{I}\right)^{-1}\mathcal{K}_{ok}$. Therefore, our CGP objective becomes
\begin{equation} \label{eq:maximize full}
\begin{aligned}
\max_\theta \Bigg\{ \hat{\mathfrak{L}}(\theta) &\triangleq  
\alpha \Big(-\frac{1}{n}\sum_{i=1}^n \Big[\left(\boldsymbol{\nu}_a-\mathbf{m}_q^i\right)^\top \boldsymbol{\Sigma}_{q}^{-1}\left(\boldsymbol{\nu}_a-\mathbf{m}_{q}^i\right)\Big]  -\log \mathrm{det}(\boldsymbol{\Sigma}_{q})\\
- \frac{1}{n}&\sum_{i=1}^n \Big[\left(\mathbf{z}_k-\mathbf{m}_k^i\right)^\top \boldsymbol{\Sigma}_{k}^{-1}\left(\mathbf{z}_k-\mathbf{m}_k^i\right)\Big]  - \log \mathrm{det}(\boldsymbol{\Sigma}_{k}) \Big) - \mathrm{loss}(\boldsymbol{\nu}_a) \Bigg\} \ .
\end{aligned}
\end{equation}

\section{Analytic Form of CGPT's Predictive Variance from Section \ref{sec:learning kernel}} \label{appx:B}
In Eq.~\eqref{eq: exp closed form}, we have derived the expectation $\mathbb{E}[\mathbf{z}_q \mid \mathbf{z}_k]$ of the CGP model, which then can be modeled as the predictive mean of CGPT in equation \eqref{eq:cgp_attention}. To perform uncertainty calibration, we need to further derive the predictive variance of $\mathbf{z}_q \mid \mathbf{z}_k$. We have the following identity:
\begin{eqnarray} \label{eq: variance formula}
\mathbb{V}[\mathbf{z}_q \mid \mathbf{z}_k] &=& \mathbb{E}\left[\mathbf{z}_q\mathbf{z}_q^\top\mid \mathbf{z}_k\right] -  \mathbb{E}\left[\mathbf{z}_q|\mathbf{z}_k]\cdot \mathbb{E}[\mathbf{z}_q|\mathbf{z}_k\right]^\top,
\end{eqnarray}
where $\mathbb{E}[\mathbf{z}_q|\mathbf{z}_k]$ is the predictive mean given in \eqref{eq: exp closed form} and $\mathbb{E}\left[\mathbf{z}_q\mathbf{z}_q^\top\mid \mathbf{z}_k\right]$ is given by the following integral, 
\begin{eqnarray} 
\label{eq:temp1}
\mathbb{E}\Big[\mathbf{z}_q\mathbf{z}_q^\top\mid \mathbf{z}_k\Big] &=& \int_{\mathbf{z}_q} \mathbf{z}_q\mathbf{z}_q^\top
\Bigg(\int_{\mathbf{z}_o}p\left(\mathbf{z}_q\mathbf{z}_q^\top\mid \mathbf{z}_o\right)p(\mathbf{z}_o\mid\mathbf{z}_k)d\mathbf{z}_o\Bigg)\mathrm{d}\mathbf{z}_q \\
&=& \int_{\mathbf{z}_0}\int_{\mathbf{z}_q} \mathbf{z}_q\mathbf{z}_q^\top p\left(\mathbf{z}_q\mathbf{z}_q^\top\mid\mathbf{z}_o\right)p(\mathbf{z}_o\mid\mathbf{z}_k)\mathrm{d}\mathbf{z}_o\mathrm{d}\mathbf{z}_q\\
&=&\int_{\mathbf{z}_o}\mathbb{E}\Big[\mathbf{z}_q\mathbf{z}_q^\top\mid\mathbf{z}_o\Big]p(\mathbf{z}_o\mid\mathbf{z}_k)\mathrm{d}\mathbf{z}_o = \mathbb{E}\Big[\mathbb{E}\Big[\mathbf{z}_q\mathbf{z}_q^\top\mid\mathbf{z}_o\Big]\mid\mathbf{z}_k\Big].
\end{eqnarray}
By the canonical representation of GP, we have 
\begin{eqnarray}
\mathbf{z}_q\mid\mathbf{z}_o &\sim& \mathcal{N}\Big(\mathcal{K}_{qo}(\mathcal{K}_o+\sigma^2\mathbf{I})^{-1}\mathbf{z}_o, \mathcal{K}_q-\mathcal{K}_{qo}(\mathcal{K}_o+\sigma^2\mathbf{I})^{-1}\mathcal{K}_{oq}\Big).
\end{eqnarray}    
Thus, using the identity $\mathbb{E}(\mathbf{x}\mathbf{x}^\top) \ = \  \boldsymbol{\Sigma} \ + \ \mathbf{m}\mathbf{m}^\top \text{ for } \mathbf{x} \sim \mathcal{N}(\mathbf{m}, \boldsymbol{\Sigma})$ we have,
\begin{eqnarray}
\mathbb{E}\Big[\mathbf{z}_q\mathbf{z}_q^\top\mid \mathbf{z}_o\Big]
&=& \mathcal{K}_q-\mathcal{K}_{qo}(\mathcal{K}_o+\sigma^2 \mathbf{I})^{-1}\mathcal{K}_{oq} + \mathcal{K}_{qo}(\mathcal{K}_o+\sigma^2\mathbf{I})^{-1}\mathbf{z}_o\mathbf{z}_o^\top (\mathcal{K}_o+\sigma^2\mathbf{I})^{-1} \mathcal{K}_{oq} \ .
\end{eqnarray}
Next, taking the expectation of $\mathbb{E}[\mathbf{z}_q\mathbf{z}_q^\top|\mathbf{z}_o]$ with respect to $\mathbf{z}_o\mid \mathbf{z}_k$, gives
\begin{eqnarray} \label{eq:temp2}
\hspace{-9mm}\mathbb{E}_{\mathbf{z}_o\mid\mathbf{z}_k}\Big[\mathbb{E}\left[\mathbf{z}_q\mathbf{z}_q^\top\mid \mathbf{z}_o\right]\Big] &=& \mathcal{K}_q-\mathcal{K}_{qo}(\mathcal{K}_o+\sigma^2 \mathbf{I})^{-1}\mathcal{K}_{oq}  
\ +\ \mathcal{K}_{qo}(\mathcal{K}_o+\sigma^2 \mathbf{I})^{-1}\mathbb{E}\left[\mathbf{z}_o\mathbf{z}_o^\top\mid \mathbf{z}_k\right] (\mathcal{K}_o+\sigma^2\mathbf{I})^{-1} \mathcal{K}_{oq}.
\end{eqnarray}
Note that $\mathbf{z}_o\mid\mathbf{z}_k \sim \mathcal{N}(\mathcal{K}_{ok}(\mathcal{K}_k+\sigma^2\mathbf{I})^{-1}\mathbf{z}_k, \mathcal{K}_o - \mathcal{K}_{ok}(\mathcal{K}_k+\sigma^2\mathbf{I})^{-1}\mathcal{K}_{ko})$ due to the canonical GP representation. Thus, we have
\begin{eqnarray} 
\label{eq:temp3}
\mathbb{E}\Big[\mathbf{z}_o\mathbf{z}_o^\top\mid \mathbf{z}_k\Big] &=& \mathcal{K}_o - \mathcal{K}_{ok}(\mathcal{K}_k+\sigma^2 \mathbf{I})^{-1}\mathcal{K}_{ok} + \mathcal{K}_{ok}(\mathcal{K}_k+\sigma^2\mathbf{I})^{-1}\mathbf{z}_k \mathbf{z}_k^\top (\mathcal{K}_k+\sigma^2\mathbf{I})^{-1} \mathcal{K}_{ko} \ .
\end{eqnarray}
Hence, we can obtain the closed form of the predictive variance $\mathbb{V}[\mathbf{z}_q \mid \mathbf{z}_k]$ by putting together Eq.~\eqref{eq: variance formula}, Eq.~\eqref{eq: exp closed form}, Eq.~\eqref{eq:temp1}, Eq.~\eqref{eq:temp2} and Eq.~\eqref{eq:temp3}. This consequently allows us to perform uncertainty calibration for the CGP-based attention unit's output analytically.

\section{Sparse CGPT Predictive Mean} \label{sec: predictive mean}
We need to find the predictive mean,
\begin{eqnarray}
\hspace{-13mm}\mathbb{E}\Big[\mathbf{z}_q \mid \mathbf{z}_k\Big] &=& \mathbb{E}_{\mathbf{z}_o \sim p(\mathbf{z}_o \mid \mathbf{z}_k)}\Bigg[\mathbb{E}\Big[\mathbf{z}_q \mid \mathbf{z}_o\Big] \mid \mathbf{z}_k\Bigg].\label{eq:pred}
\end{eqnarray}
The distribution $\mathbf{z}_q \mid \mathbf{z}_o$ can be approximated using sparse GP techniques, such as DTC,
\begin{eqnarray}
    p(\mathbf{z}_q \mid \mathbf{z}_o) &=& \int_{\mathbf{z}_m} p(\mathbf{z}_q \mid \mathbf{z}_m) p(\mathbf{z}_m \mid \mathbf{z}_o) \mathrm{d} \mathbf{z}_m,
\end{eqnarray}
where $p(\mathbf{z}_m \mid \mathbf{z}_o)$ is the inducing posterior and has the form
\begin{equation}
    \begin{aligned}
    p(\mathbf{z}_m \mid \mathbf{z}_o) &=&\mathbb{N} \Big (\mathbf{z}_m \mid 
    \frac{1}{\sigma^2} \mathcal{K}_{mm}\Big (\mathcal{K}_{mm} + \frac{1}{\sigma^2} \mathcal{K}_{mo}\mathcal{K}_{om}\Big )^{-1}\mathcal{K}_{mo}\mathbf{z}_o,
    & \mathcal{K}_{mm} \Big (\mathcal{K}_{mm} + \frac{1}{\sigma^2} \mathcal{K}_{mo}\mathcal{K}_{om}\Big )^{-1}\mathcal{K}_{mm} \Big ).
    \end{aligned}
\end{equation}
The distribution $p(\mathbf{z}_q \mid \mathbf{z}_m)$ has the form $\mathbb{N}(\mathbf{z}_q \mid \mathcal{K}_{qm}\mathcal{K}_{mm}^{-1}\mathbf{z}_m, \sigma^2 \mathbf{I})$. Therefore, with some algebras, we can calculate the expectation of $\mathbf{z}_q\mid\mathbf{z}_o$,
\begin{equation}
    \begin{aligned}
        \mathbb{E}(\mathbf{z}_q\mid\mathbf{z}_o) &=  \int_{\mathbf{z}_q}\mathbf{z}_q\int_{\mathbf{z}_m} p(\mathbf{z}_q \mid \mathbf{z}_m) p(\mathbf{z}_m \mid \mathbf{z}_o) \mathrm{d} \mathbf{z}_m \mathrm{d} \mathbf{z}_q \\
        &=\int_{\mathbf{z}_m}\Big(\int_{\mathbf{z}_q} \mathbf{z}_q p(\mathbf{z}_q \mid \mathbf{z}_m) \mathrm{d} \mathbf{z}_q\Big) p(\mathbf{z}_m \mid \mathbf{z}_o) \mathrm{d} \mathbf{z}_m \\
        &= \int_{\mathbf{z}_m} \mathbf{z}_q \mathcal{K}_{qm}\mathcal{K}_{mm}^{-1}\mathbf{z}_m p(\mathbf{z}_m \mid \mathbf{z}_o) \mathrm{d} \mathbf{z}_m  \\
        &=\frac{1}{\sigma^2} \mathcal{K}_{qm}\mathcal{K}_{mm}^{-1} \mathcal{K}_{mm}\Big (\mathcal{K}_{mm} + \frac{1}{\sigma^2} \mathcal{K}_{mo}\mathcal{K}_{om}\Big )^{-1}\mathcal{K}_{mo}\mathbf{z}_o \\
        &= \frac{1}{\sigma^2} \mathcal{K}_{qm}\Big (\mathcal{K}_{mm} + \frac{1}{\sigma^2} \mathcal{K}_{mo}\mathcal{K}_{om}\Big )^{-1}\mathcal{K}_{mo}\mathbf{z}_o.
    \end{aligned}
\end{equation}
    
In a similar fashion, we can find the expectation of $\mathbf{z}_o\mid\mathbf{z}_k$ 
\begin{eqnarray}
    \mathbb{E}(\mathbf{z}_o\mid\mathbf{z}_k) &=&  \frac{1}{\sigma^2} \mathcal{K}_{ol}\Big (\mathcal{K}_{ll} + \frac{1}{\sigma^2} \mathcal{K}_{lk}\mathcal{K}_{kl}\Big )^{-1}\mathcal{K}_{lk}\mathbf{z}_k.
\end{eqnarray}
Since $\mathbf{z}_q\mid\mathbf{z}_o$ and $\mathbf{z}_o\mid\mathbf{z}_k$ are Gaussians, we can analytically calculate
\begin{equation}
    \begin{aligned}
        \mathbb{E}(\mathbf{z}_q\mid\mathbf{z}_k) &= \frac{1}{\sigma^2} \mathcal{K}_{qm}\Big (\mathcal{K}_{mm} + \frac{1}{\sigma^2} \mathcal{K}_{mo}\mathcal{K}_{om}\Big )^{-1}\mathcal{K}_{mo} \cdot \frac{1}{\sigma^2} \mathcal{K}_{ol}\Big (\mathcal{K}_{ll} + \frac{1}{\sigma^2} \mathcal{K}_{lk}\mathcal{K}_{kl}\Big )^{-1}\mathcal{K}_{lk}\mathbf{z}_k \\
        &= \frac{1}{\sigma^4} \mathcal{K}_{qm}\Big (\mathcal{K}_{mm} + \frac{1}{\sigma^2} \mathcal{K}_{mo}\mathcal{K}_{om}\Big )^{-1}\mathcal{K}_{mo} \mathcal{K}_{ol}\Big (\mathcal{K}_{ll} + \frac{1}{\sigma^2} \mathcal{K}_{lk}\mathcal{K}_{kl}\Big )^{-1}\mathcal{K}_{lk}\mathbf{z}_k. \\
    \end{aligned}
\end{equation}

\section{Sparse CGPT Predictive Variance}
The variance of $\mathbf{z}_q\mid \mathbf{z}_k$ is given by
\begin{align} \label{eq: predictive variance}
    \mathbb{V}[\mathbf{z}_q\mid \mathbf{z}_k] = \mathbb{E}[\mathbf{z}_q\mathbf{z}_q^\top\mid \mathbf{z}_k] - \mathbb{E}[\mathbf{z}_q\mid \mathbf{z}_k]\mathbb{E}[\mathbf{z}_q\mid \mathbf{z}_k]^\top.
\end{align}
where $\mathbb{E}[\mathbf{z}_q\mid \mathbf{z}_k]$ is the predictive mean in Section \ref{sec: predictive mean}. The expectation of $\mathbf{z}_q\mathbf{z}_q\top|z_k$ is given by
\begin{align} \label{eq: quad zq|zk}
    \mathbb{E}[\mathbf{z}_q\mathbf{z}_q^\top|\mathbf{z}_k] = \mathbb{E}_{\mathbf{z}_o \sim p(\mathbf{z}_o \mid \mathbf{z}_k)}\Bigg[\mathbb{E}\Big[\mathbf{z}_q\mathbf{z}_q^\top \mid \mathbf{z}_o\Big] \mid \mathbf{z}_k\Bigg].
\end{align}
Consider,
\begin{align}   \mathbb{E}\Big[\mathbf{z}_q\mathbf{z}_q^\top \mid \mathbf{z}_o\Big] &=\int_{\mathbf{z}_q} \mathbf{z}_q\mathbf{z}_q^\top p(\mathbf{z}_q|\mathbf{z}_o)\mathrm{d}\mathbf{z}_q = \int_{\mathbf{z}_q} \mathbf{z}_q\mathbf{z}_q^\top \int_{\mathbf{z}_m}p(\mathbf{z}_q|\mathbf{z}_m)p(\mathbf{z}_m|\mathbf{z}_o) \mathrm{d}\mathbf{z}_m\mathrm{d}\mathbf{z}_q\\
&= \int_{\mathbf{z}_m}\Big(\int_{\mathbf{z}_q} \mathbf{z}_q\mathbf{z}_q^\top p(\mathbf{z}_q|\mathbf{z}_m)\mathrm{d}\mathbf{z}_q \Big)p(\mathbf{z}_m|\mathbf{z}_o)\mathrm{d}\mathbf{z}_m = \int_{\mathbf{z}_m}\mathbb{E}[\mathbf{z}_q\mathbf{z}_q^\top\mid\mathbf{z}_m]p(\mathbf{z}_m|\mathbf{z}_o)\mathrm{d}\mathbf{z}_m.
\end{align}

Since $p(\mathbb{E}[\mathbf{z}_q\mid\mathbf{z}_m])=\mathbb{N}(\mathbf{z}_q \mid \mathcal{K}_{qm}\mathcal{K}_{mm}^{-1}\mathbf{z}_m, \sigma^2 \mathbf{I})$, we have,
\begin{align*}
    \mathbb{E}[\mathbf{z}_q\mathbf{z}_q^\top\mid\mathbf{z}_m] &= \sigma^2\mathbf{I} + \mathcal{K}_{qm}\mathcal{K}_{mm}^{-1}\mathbf{z}_m \mathbf{z}_m^\top\mathcal{K}_{mm}^{-1}\mathcal{K}_{mq}.
\end{align*}
Therefore,
\begin{equation}
    \begin{aligned} \label{eq:quad zq|zo}
    \mathbb{E}\Big[\mathbf{z}_q\mathbf{z}_q^\top \mid \mathbf{z}_o\Big] &= \int_{\mathbf{z}_m}(\sigma^2\mathbf{I} + \mathcal{K}_{qm}\mathcal{K}_{mm}^{-1}\mathbf{z}_m \mathbf{z}_m^\top\mathcal{K}_{mm}^{-1}\mathcal{K}_{mq})p(\mathbf{z}_m|\mathbf{z}_o)\mathrm{d}\mathbf{z}_m\\
    &= \sigma^2\mathbf{I} + \mathcal{K}_{qm}\mathcal{K}_{mm}^{-1} \int_{\mathbf{z}_m} \mathbf{z}_m \mathbf{z}_m^\top p(\mathbf{z}_m|\mathbf{z}_o)\mathrm{d}\mathbf{z}_m \mathcal{K}_{mm}^{-1}\mathcal{K}_{mq}\\
    &= \sigma^2\mathbf{I} + \mathcal{K}_{qm}\mathcal{K}_{mm}^{-1} \mathbb{E}[\mathbf{z}_m\mathbf{z}_m^\top\mid \mathbf{z}_o]\mathcal{K}_{mm}^{-1}\mathcal{K}_{mq}.
\end{aligned}
\end{equation}

Since $p(\mathbf{z}_m\mid \mathbf{z}_o)=\mathbb{N} \Big (\mathbf{z}_m \mid 
\frac{1}{\sigma^2} \mathcal{K}_{mm}\Big (\mathcal{K}_{mm} + \frac{1}{\sigma^2} \mathcal{K}_{mo}\mathcal{K}_{om}\Big )^{-1}\mathcal{K}_{mo}\mathbf{z}_o,
\mathcal{K}_{mm} \Big (\mathcal{K}_{mm} + \frac{1}{\sigma^2} \mathcal{K}_{mo}\mathcal{K}_{om}\Big )^{-1}\mathcal{K}_{mm} \Big )$, we have the following
\begin{equation} \label{eq: quad zm|zo}
    \begin{aligned}
    \mathbb{E}[\mathbf{z}_m\mathbf{z}_m^\top\mid \mathbf{z}_o] &= \mathcal{K}_{mm} \Big (\mathcal{K}_{mm} + \frac{1}{\sigma^2} \mathcal{K}_{mo}\mathcal{K}_{om}\Big )^{-1}\mathcal{K}_{mm} + \frac{1}{\sigma^4} \mathcal{K}_{mm}\Big (\mathcal{K}_{mm} + \frac{1}{\sigma^2} \mathcal{K}_{mo}\mathcal{K}_{om}\Big )^{-1}\mathcal{K}_{mo}\mathbf{z}_o \times \\
    &\mathbf{z}_o^\top \mathcal{K}_{om}\Big (\mathcal{K}_{mm} + \frac{1}{\sigma^2} \mathcal{K}_{mo}\mathcal{K}_{om}\Big )^{-1}\mathcal{K}_{mm}.
\end{aligned}
\end{equation}
Combining Eq \eqref{eq:quad zq|zo} and \eqref{eq: quad zm|zo}, we have
\begin{equation} \label{eq: quad zq|zo full}
    \begin{aligned}
    \mathbb{E}\Big[\mathbf{z}_q\mathbf{z}_q^\top \mid \mathbf{z}_o\Big] &= \sigma^2\mathbf{I} + \mathcal{K}_{qm}\Big[\Big (\mathcal{K}_{mm} + \frac{1}{\sigma^2} \mathcal{K}_{mo}\mathcal{K}_{om}\Big )^{-1} + \frac{1}{\sigma^4} \Big (\mathcal{K}_{mm} + \frac{1}{\sigma^2} \mathcal{K}_{mo}\mathcal{K}_{om}\Big )^{-1} \mathcal{K}_{mo} \mathbf{z}_o \times \\
    &\mathbf{z}_o^\top\mathcal{K}_{om}\Big (\mathcal{K}_{mm} + \frac{1}{\sigma^2} \mathcal{K}_{mo}\mathcal{K}_{om}\Big )^{-1} \Big] \mathcal{K}_{mq}. 
\end{aligned}
\end{equation}

Plug Eq \eqref{eq: quad zq|zo full} to Eq \eqref{eq: quad zq|zk}, we have
\begin{equation} \label{eq: quad zq|zk 2}
    \begin{aligned}
    \mathbb{E}[\mathbf{z}_q\mathbf{z}_q^\top|\mathbf{z}_k] 
    &= \sigma^2\mathbf{I} + \mathcal{K}_{qm}\Big[\Big (\mathcal{K}_{mm} + \frac{1}{\sigma^2} \mathcal{K}_{mo}\mathcal{K}_{om}\Big )^{-1} + \frac{1}{\sigma^4} \Big (\mathcal{K}_{mm} + \frac{1}{\sigma^2} \mathcal{K}_{mo}\mathcal{K}_{om}\Big )^{-1} \mathcal{K}_{mo} \times \\
    &\Big(\int_{\mathbf{z}_o}\mathbf{z}_o\mathbf{z}_o^\top p(\mathbf{z}_o\mid
    \mathbf{z}_k)\mathrm{d}\mathbf{z}_o\Big)\mathcal{K}_{om}\Big (\mathcal{K}_{mm} + \frac{1}{\sigma^2} \mathcal{K}_{mo}\mathcal{K}_{om}\Big )^{-1} \Big] \mathcal{K}_{mq}. 
\end{aligned}
\end{equation}

In a similar manner, we can calculate the integral
\begin{equation} \label{eq: quad zo|zk full}
    \begin{aligned}
    &\int_{\mathbf{z}_o}\mathbf{z}_o\mathbf{z}_o^\top p(\mathbf{z}_o\mid
    \mathbf{z}_k)\mathrm{d}\mathbf{z}_o =  \mathbb{E}[\mathbf{z}_o\mathbf{z}_o^\top|\mathbf{z}_k] \\
    &=\sigma^2\mathbf{I} + \mathcal{K}_{ol}\Big[\Big (\mathcal{K}_{ll} + \frac{1}{\sigma^2} \mathcal{K}_{lk}\mathcal{K}_{kl}\Big )^{-1} + \frac{1}{\sigma^4} \Big (\mathcal{K}_{ll} + \frac{1}{\sigma^2} \mathcal{K}_{lk}\mathcal{K}_{kl}\Big )^{-1} \mathcal{K}_{lk} \mathbf{z}_k\mathbf{z}_k^\top\mathcal{K}_{kl}\Big (\mathcal{K}_{ll} + \frac{1}{\sigma^2} \mathcal{K}_{lk}\mathcal{K}_{kl}\Big )^{-1} \Big] \mathcal{K}_{lo}. 
\end{aligned}
\end{equation}

From equation \eqref{eq: quad zo|zk full}, \eqref{eq: quad zq|zk 2} and \eqref{eq: predictive variance}, we have the full predictive varicance of sparse CGPT.

\section{Derivation of Sparse CGP Loss Function} \label{sec: SCGP objective}
The objective function of Sparse CGP is given by
\begin{equation} \label{eq:maximize}
    \begin{aligned}
        \min_\theta \Big\{ \mathfrak{L}(\theta) &\triangleq \mathrm{loss}(\boldsymbol{\nu}_a) - \alpha \cdot \Big(\log \mathbb{E}_{\mathbf{z}_o}\big[p(\mathbf{z}_q = \boldsymbol{\nu}_a \mid \mathbf{z}_o)\big] \ +\ \log \mathbb{E}_{\mathbf{z}_o}\big[p(\mathbf{z}_k \mid \mathbf{z}_o)\big]\Big) \Big\}.
    \end{aligned}
\end{equation}

We will optimize the lower bound of $\log p(\mathbf{z}_q = \boldsymbol{\nu}_a\mid \mathbf{z}_o)$ and $\log p(\mathbf{z}_k \mid \mathbf{z}_o)$. Consider $p(\mathbf{z}_q = \boldsymbol{\nu}_a \mid \mathbf{z}_o)$, we have
\begin{equation} \label{eq: log q}
    \begin{aligned}
        \log \mathbb{E}_{\mathbf{z}_o} \Big[ p(\mathbf{z}_q=\boldsymbol{\nu}_a \mid \mathbf{z}_o)\Big] 
        &\geq \mathbb{E}_{\mathbf{z}_o}\log \Big[ p(\mathbf{z}_q =\boldsymbol{\nu}_a\mid \mathbf{z}_o)\Big] = \mathbb{E}_{\mathbf{z}_o} \Big[\log \mathbb{E}_{\mathbf{z}_m\mid\mathbf{z}_o}[p(\mathbf{z}_q=\boldsymbol{\nu}_a\mid \mathbf{z}_m)] \Big] \\&\geq \mathbb{E}_{\mathbf{z}_o} \Big[ \mathbb{E}_{\mathbf{z}_m\mid\mathbf{z}_o}\Big[\log [p(\mathbf{z}_q=\boldsymbol{\nu}_a\mid \mathbf{z}_m)] \Big] \Big ],
    \end{aligned}
\end{equation}
where we use Jensen's inequality to lower bound the log expectation. Since $\mathbf{z}_q\mid \mathbf{z}_m \sim \mathbb{N}(\mathcal{K}_{qm}\mathcal{K}_{mm}^{-1}\mathbf{z}_m, \sigma^2\mathbf{I})$, we have the following
\begin{eqnarray} \label{eq: bound}
    \mathbb{E}_{\mathbf{z}_m\mid\mathbf{z}_o}\Big[\log p(\mathbf{z}_q=\boldsymbol{\nu}_a\mid\mathbf{z}_m) \Big] &= -\frac{1}{2\sigma^2}\mathbb{E}_{\mathbf{z}_m\mid\mathbf{z}_o}  \Big[ ||\boldsymbol{\nu}_a-\mathcal{K}_{qm}\mathcal{K}_{mm}^{-1}\mathbf{z}_m||^2 \Big] -n\log 2\pi - \frac{1}{2\sigma}.
\end{eqnarray}
We have the following identity: If $X\sim\mathbb{N}(\mu, \Sigma)$ and $A$ is a symmetric matrix, then $\mathbb{E}[X^\top A X] = \mu^\top A \mu + \text{trace}(A\Sigma)$. Since $\mathbf{z}_m\mid\mathbf{z}_o \sim \mathbb{N}\Big( \frac{1}{\sigma^2}\mathcal{K}_{mm}\Big(\mathcal{K}_{mm}+\frac{1}{\sigma^2}\mathbf{I}\Big)^{-1}\mathcal{K}_{mo}\mathbf{z}_o, \mathcal{K}_{mm}\Big(\mathcal{K}_{mm}+\frac{1}{\sigma^2}\mathbf{I}\Big)^{-1}\mathcal{K}_{mm}\Big)$, following the above identity with $X=\boldsymbol{\nu}_a-\mathcal{K}_{qm}\mathcal{K}_{mm}^{-1}\mathbf{z}_m$ and $A=\mathbf{I}$, we have
\begin{align} \label{eq: analytic}
    \mathbb{E}_{\mathbf{z}_m\mid\mathbf{z}_o}  \Big[ ||\boldsymbol{\nu}_a-\mathcal{K}_{qm}\mathcal{K}_{mm}^{-1}\mathbf{z}_m||^2 \Big] = \Big|\Big|\boldsymbol{\nu}_a - \frac{1}{\sigma^2}\mathcal{K}_{qm}\Big(\mathcal{K}_{mm}+\frac{1}{\sigma^2}\mathbf{I}\Big)^{-1}\mathcal{K}_{mo}\mathbf{z}_o\Big|\Big|^2 + \text{trace}\Big[\mathcal{K}_{qm}\Big(\mathcal{K}_{mm}+\frac{1}{\sigma^2}\mathbf{I}\Big)^{-1}\mathcal{K}_{mq}\Big].
\end{align}

Taking the expectation over $\mathbf{z}_o \sim \mathbb{N}(\mathbf{0}, \mathcal{K}_{oo})$, we have
\begin{equation} \label{eq: final lower bound}
    \begin{aligned}
        \mathbb{E}_{\mathbf{z}_o} \Big[ \mathbb{E}_{\mathbf{z}_m\mid\mathbf{z}_o}  \Big[ ||\boldsymbol{\nu}_a-\mathcal{K}_{qm}\mathcal{K}_{mm}^{-1}\mathbf{z}_m||^2 \Big]\Big ] &=
  ||\boldsymbol{\nu}_a||^2 + \frac{1}{\sigma^4}\text{trace}\Big[ 
\mathcal{K}_{om}\Big(\mathcal{K}_{mm}+\frac{1}{\sigma^2}\mathbf{I}\Big)^{-1}\mathcal{K}_{mq} \mathcal{K}_{qm}\Big(\mathcal{K}_{mm}+\frac{1}{\sigma^2}\mathbf{I}\Big)^{-1}\mathcal{K}_{mo}\mathcal{K}_{oo}\Big] +\\
  &\text{trace}\Big[\mathcal{K}_{qm}\Big(\mathcal{K}_{mm}+\frac{1}{\sigma^2}\mathbf{I}\Big)^{-1}\mathcal{K}_{mq}\Big].
    \end{aligned}
\end{equation}


Combining \eqref{eq: log q}, \eqref{eq: bound}, \eqref{eq: analytic} and \eqref{eq: final lower bound}, we have the closed form lower bound for $\log p(\mathbf{z}_q \mid \mathbf{z}_o)$. Using similar argument, we also obtain the lower bound for $\log p(\mathbf{z}_k \mid \mathbf{z}_o)$.

\section{Additional Experiment Results} \label{app:C}
    \subsection{Out-of-Distribution Calibration}
       \textbf{CIFAR100-C.} 
      This section expands on our previous empirical comparison between SGPA and CGPT. Previously, we have shown that CGPT has comparable performances with SGPA in terms of accuracy (MCC) and has much better uncertainty calibration on the CIFAR10 dataset (see Table~\ref{tab:in-distribution} and Table~\ref{tab:OOD CIFAR}). In addition, to compare the robust performance of CGPT and SGPA on larger scale OOD learning scenarios, we also use the corrupted CIFAR100-C dataset. Similar to the CIFAR10-C dataset, the CIFAR100-C dataset also contains corrupted images from CIFAR100, which can be divided into $19$ types of distortion belonging to $4$ distortion categories: Noise, Blur, Weather and Digital. For each method, we calculate the mean performance metrics over the distortion types in each distortion category. The results in Table \ref{tab:OOD CIFAR100 appx} shows that while CGPT has comparable accuracy with the SGPA baseline, the calibration capacity of CGPT is much better than SGPA  with lower NLL, MCE and ECE across all types of distortion.

      \par \textbf{CIFAR10-C.} We provide additional results for CGPT on CIFAR10-C. Beside the results for CGPT with the value $\alpha$ gradually increases from $0.5$ to $1.0$ during training as in Table~\ref{tab:OOD CIFAR}, we also train another CGPT with fixed value of $\alpha=0.7$. We found that CGPT in this setting can help achieve better accuracy and calibration results, which are shown in Table \ref{tab:OOD CIFAR10 appendix}.    
        \begin{table*}[t]
        \centering
        \caption{Test Accuracy and other calibration metrics achieved by our CGPT model with 2 different settings of $\alpha$ on CIFAR10-C dataset. For each of the $4$ distortion categories, we report the mean metrics over all distortion types in the category. And for each reported result, we run with 3 random seeds and report mean and standard deviation.}
        \vspace{0.5em}
        \resizebox{17cm}{!}{
        \begin{tabular}{|l l l l l l l|} 
            \toprule
            \textbf{Metric} & \textbf{Model} & \bf{Noise} & \bf{Blur} & \bf{Weather} & \bf{Digital} & \bf{Avg.}\\
            \midrule
            
            \multirow{2}{*}{Acc $\uparrow$} & SGPA & 50.803 $\pm$ 0.447 & \textbf{59.264 $\pm$ 0.915} & \textbf{64.148 $\pm$ 0.472} & \textbf{63.028 $\pm$ 0.334} & \textbf{59.722 $\pm$ 0.323}\\ 
            & CGPT ($\alpha=0.5 \rightarrow 1.0$)& \textbf{55.177 $\pm$ 0.953} & 56.412 $\pm$ 1.506 & 61.515 $\pm$ 0.703 & 60.373 $\pm$ 0.123 & 58.591 $\pm$ 0.664\\
            & CGPT ($\alpha=0.7$)& 54.110 $\pm$ 0.298 & 58.056 $\pm$ 0.233 & 61.655 $\pm$ 0.348 & 61.029 $\pm$ 0.258 & 58.971 $\pm$ 0.111 \\
            \midrule

             \multirow{2}{*}{NLL $\downarrow$} & SGPA & 3.464 $\pm$ 0.423 & 2.551 $\pm$ 0.091 & 2.137 $\pm$ 0.162 & 2.298 $\pm$ 0.045 & 2.626 $\pm$ 0.202\\ 
            & CGPT ($\alpha=0.5 \rightarrow 1.0$) & 1.688 $\pm$ 0.033 & 1.565 $\pm$ 0.068 & 1.352 $\pm$ 0.049 & 1.461 $\pm$ 0.027 & 1.516 $\pm$ 0.029\\
            & CGPT ($\alpha=0.7$)& \textbf{1.670 $\pm$ 0.180} & \textbf{1.403 $\pm$ 0.131} & \textbf{1.281 $\pm$ 0.132} & \textbf{1.341 $\pm$ 0.099} & \textbf{1.414 $\pm$ 0.131}\\
            \midrule

             \multirow{2}{*}{MCE $\downarrow$} & SGPA & 0.668$\pm$ 0.009 & 0.592 $\pm$ 0.014 & 0.576 $\pm$ 0.014 & 0.575 $\pm$ 0.001 & 0.593 $\pm$ 0.002\\ 
            & CGPT ($\alpha = 0.5 \rightarrow 1.0$) & \textbf{0.360 $\pm$ 0.011} & 0.334 $\pm$ 0.013 & \textbf{0.284 $\pm$ 0.002} & \textbf{0.314 $\pm$ 0.003} & \textbf{0.324 $\pm$ 0.002}\\
            & CGPT ($\alpha=0.7$)& 0.379 $\pm$ 0.025 & \textbf{0.330 $\pm$ 0.009} & 0.299 $\pm$ 0.017 & 0.318 $\pm$ 0.000 & 0.330 $\pm$ 0.011\\
            \midrule

             \multirow{2}{*}{ECE $\downarrow$} & SGPA & 0.532 $\pm$ 0.021 & 0.488 $\pm$ 0.012 & 0.469 $\pm$ 0.003 & 0.472 $\pm$ 0.010 & 0.487 $\pm$ 0.012\\ 
            & CGPT ($\alpha=0.5 \rightarrow 1.0$) & \textbf{0.226 $\pm$ 0.012} & \textbf{0.202 $\pm$ 0.007} & \textbf{0.159 $\pm$ 0.004} & \textbf{0.183 $\pm$ 0.003} & \textbf{0.192 $\pm$ 0.001}\\
            & CGPT ($\alpha=0.7$)& 0.241 $\pm$ 0.021 & 0.199 $\pm$ 0.001 & 0.169 $\pm$ 0.013 & 0.180 $\pm$ 0.003 & 0.195 $\pm$ 0.007\\
            \bottomrule
        \end{tabular}}
        \label{tab:OOD CIFAR10 appendix}
    \end{table*}

    \begin{table*}[t]
        \centering
        \caption{Test Accuracy and other calibration metrics achieved by our CGPT model on CIFAR100-C dataset under the OOD setting. For each of the $4$ distortion categories, we report the mean metrics over all distortion types in the category. And for each reported result, we run with 3 random seeds and report mean and standard deviation. We again observe that CGPT attains better calibration metrics than SGPA across all cases.}
        \vspace{0.5em}
         \resizebox{16cm}{!}{
        \begin{tabular}{|l l l l l l l|} 
            \toprule
            \textbf{Metric} & \textbf{Model} & \bf{Noise} & \bf{Blur} & \bf{Weather} & \bf{Digital} & \bf{Avg.}\\
            \midrule
            
            \multirow{2}{*}{Acc $\uparrow$} & SGPA & \textbf{23.383 $\pm$ 0.308} & \textbf{36.405 $\pm$ 0.263} & \textbf{35.940 $\pm$ 0.120} & \textbf{35.533 $\pm$ 0.084} & \textbf{33.117 $\pm$ 0.126}\\ 
            & CGPT ($\alpha=0.7$) & 22.664 $\pm$ 0.007 & 34.488 $\pm$ 0.949 & 35.341 $\pm$ 0.375 & 34.259 $\pm$ 0.059 & 31.973 $\pm$ 0.313 \\
            \midrule

             \multirow{2}{*}{NLL $\downarrow$} & SGPA & 10.163 $\pm$ 0.583 & 6.987 $\pm$ 0.033 & 6.856 $\pm$ 0.050 & 7.284 $\pm$ 0.039 & 7.763 $\pm$ 0.161\\ 
            & CGPT ($\alpha=0.7$) & \textbf{5.600 $\pm$ 0.527} & \textbf{3.270 $\pm$ 0.360} & \textbf{3.197 $\pm$ 0.303} & \textbf{3.348 $\pm$ 0.272} & \textbf{3.797 $\pm$ 0.355}\\
            \midrule

             \multirow{2}{*}{MCE $\downarrow$} & SGPA & 0.723$\pm$ 0.008 & 0.628 $\pm$ 0.003 & 0.626 $\pm$ 0.004 & 0.637 $\pm$ 0.001 & 0.652 $\pm$ 0.004\\ 
            & CGPT ($\alpha=0.7$) & \textbf{0.633 $\pm$ 0.030} & \textbf{0.456 $\pm$ 0.068} & \textbf{0.459 $\pm$ 0.049} & \textbf{0.459 $\pm$ 0.057} & \textbf{0.497 $\pm$ 0.052}\\
            \midrule

             \multirow{2}{*}{ECE $\downarrow$} & SGPA & 0.597 $\pm$ 0.015 & 0.491 $\pm$ 0.004 & 0.492 $\pm$ 0.002 & 0.495 $\pm$ 0.001 & 0.521 $\pm$ 0.001\\ 
            & CGPT ($\alpha=0.7$) & \textbf{0.454 $\pm$ 0.038} & \textbf{0.294 $\pm$ 0.060} & \textbf{0.295 $\pm$ 0.055} & \textbf{0.289 $\pm$ 0.050} & \textbf{0.328 $\pm$ 0.050}\\
            \bottomrule
        \end{tabular}}
        \label{tab:OOD CIFAR100 appx}
    \end{table*}


    \subsection{CGPT Helps Reduce Oversmoothing in Transformers} \label{sec appendix: OVSMT}
        \par In this section, we conduct additional oversmoothing analysis similar to that in section \ref{sec:experiments} on the larger dataset CIFAR100. We compare the oversmoothing effect of SGPA and CGPT and use the settings for CIFAR100 detailed in Section \ref{sec:details image}. For CGPT, we fix $\alpha=0.7$ in the CGP objective function in the training phase. After training both CGPT and SGPA, we measured the cosine similarity between the outputs of the attention block in each layer to depict the oversmoothing effect.
        \par
\begin{figure}[t!]
                \centering
                \includegraphics[scale=0.55]{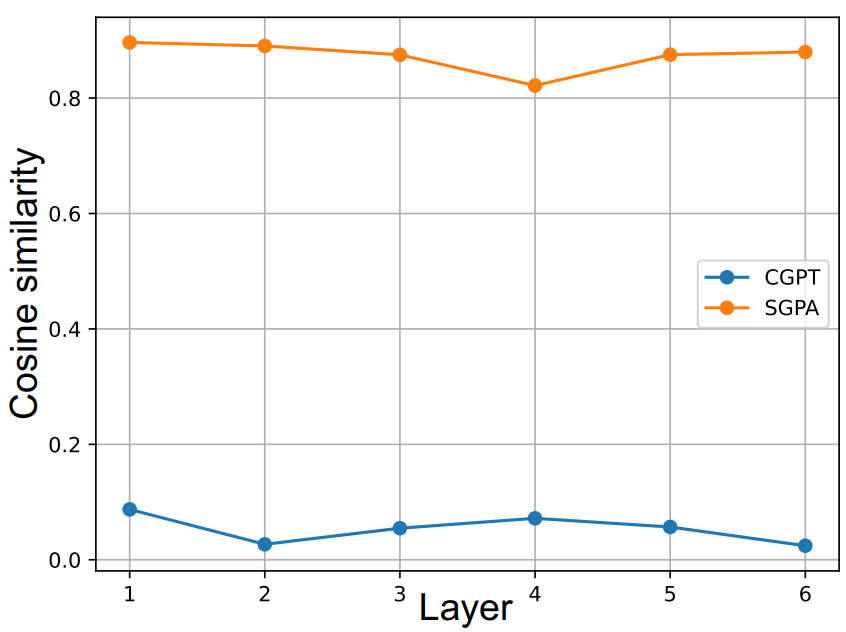}
                \vspace{-1em}
                \caption{The cosine similarity between the token representations after the attention calculation
                vs. the layer index of CGPT and SGPA on CIFAR100. CGPT is much less vulnerable to oversmoothing compared to SGPA.}
                \label{fig:over_cifar100}
                \vspace{-0.2in}
            \end{figure}

  This is visually demonstrated in Fig.~\ref{fig:over_cifar100}, which shows that as the number of attention blocks increases, the cosine similarities between the representations learned with SGPA become gradually higher. This implies that these representations will become more similar with each other as the models get deeper. On the contrary, the learned representations of CGPT have much lower cosine similarity as the model depth increases, which implies that CGPT will suffer less from oversmoothing than the SGPA.

        \begin{figure}[t!]
                \centering
                \captionsetup{font=small} 
                \includegraphics[scale=0.5]{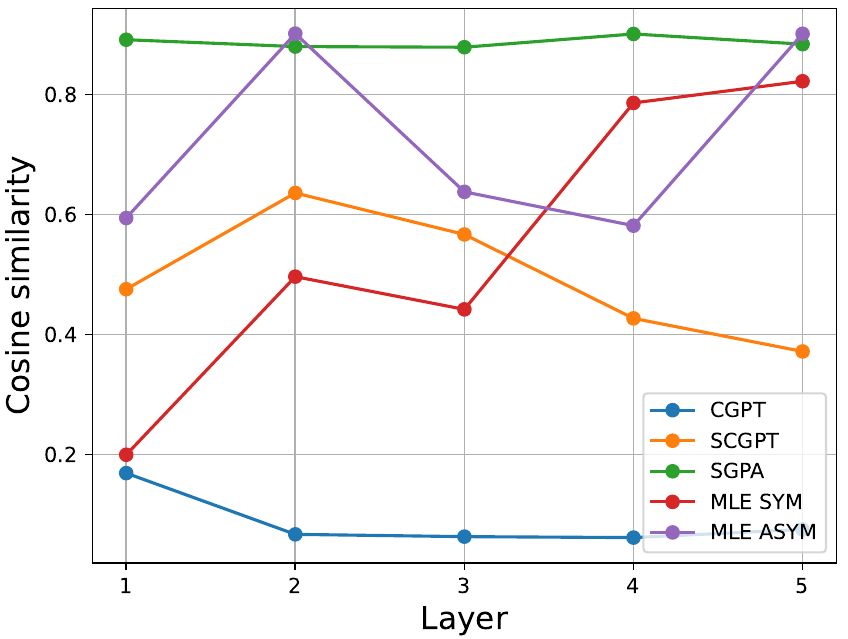}
                \vspace{-1em}
                \caption{The cosine similarity between the token representations
                vs. the layer index of CGPT and SGPA on CIFAR10. CGPT is much less vulnerable to oversmoothing compared to SGPA.}
                \label{fig:over_cifar10}
                \vspace{-0.2in}
            \end{figure}

\end{document}